\documentclass{article}

\PassOptionsToPackage{numbers, compress}{natbib}
\usepackage[preprint]{neurips_2026}

\usepackage[utf8]{inputenc} 
\usepackage[T1]{fontenc}    
\usepackage{hyperref}       
\usepackage{url}            
\usepackage{booktabs}       
\usepackage{amsfonts}       
\usepackage{nicefrac}       
\usepackage{microtype}      
\usepackage{xcolor}         
\usepackage{amsmath}
\usepackage{tabularx}
\usepackage{graphicx}
\usepackage{float}
\usepackage{multirow}
\title{C3-UniMM: Causal Cycle-Consistent Unified Multimodal Modeling via Super Alignment and Shared Decoding Space}

\author{%
  Yujie Shen\\
  Hunan University\\
  \texttt{jayshum@hnu.edu.cn} \\
  \And
  Lianlei Shan\\
  Tsinghua University\\
  \texttt{shanlianlei18@mails.ucas.edu.cn} \\
}

\begin{document}

\maketitle

\begin{abstract}
  Unified Multimodal Models aim to achieve any-to-any understanding and generation across arbitrary modalities. However, existing methods primarily rely on modeling implicit statistical correlations and lack cross-modal structural consistency constraints. This deficiency leads to profound issues, including semantic drift, poor compositional generalization, and instability under interventions. In this paper, we propose C3-UniMM, a unified multimodal modeling framework based on Causal Cycle Consistency and Super Alignment. Specifically, we introduce a Structured Latent Causal Graph (SLCG) as a shared cross-modal semantic space and design unified multimodal encoding blocks, enabling understanding and generation to be synergistically optimized within the identical causal semantic structure. Furthermore, we propose a Unified Decoding Space to enforce structural preservation and semantic invertibility during the cross-modal generation process. Theoretical analyses demonstrate that our approach significantly enhances both the invertibility and mechanism invariance of cross-modal mappings. Extensive experimental results across multiple understanding, generation, and compositional generalization tasks indicate that C3-UniMM substantially outperforms existing unified multimodal baselines.
\end{abstract}

\section{Introduction}

\begin{figure}[h]
    \centering
    \includegraphics[width=1\textwidth]{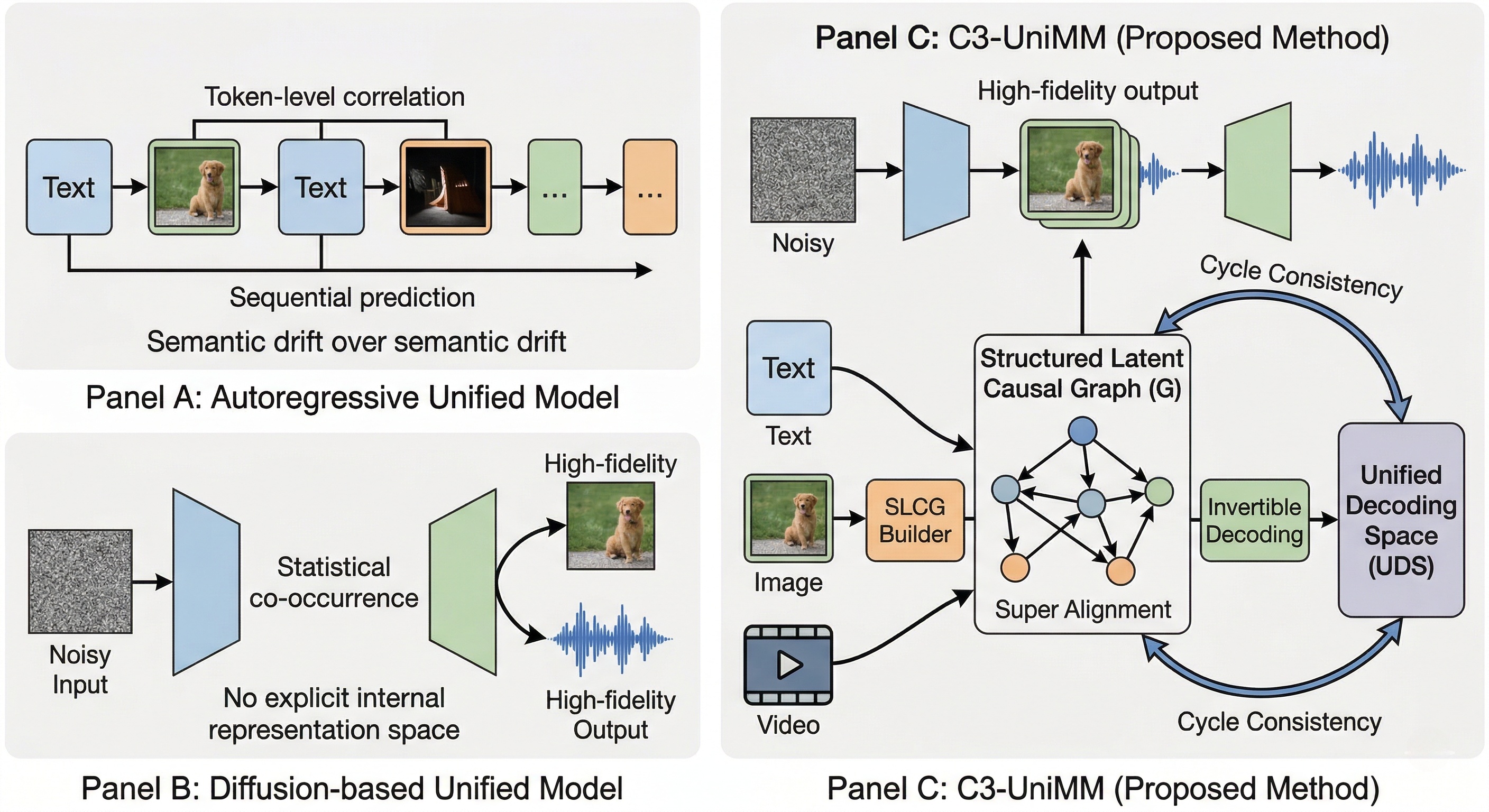}
    \caption{Comparison of different architectures. (Panel A) Autoregressive Model: Utilizes sequential token prediction and token-level correlation, which is prone to cumulative semantic drift.(Panel B) Diffusion-based Model: Relies on statistical co-occurrence to generate high-fidelity output but lacks an explicit internal representation space.(Panel C) C3-UniMM (Ours): Features a Structured Latent Causal Graph (G) and Super Alignment to integrate multimodal inputs. It utilizes a Unified Decoding Space (UDS) and Cycle Consistency to ensure structural integrity and high-fidelity generation}
    \label{fig 1}
\end{figure}

Unified multimodal modeling is emerging as a key research direction in multimodal AI. Unlike traditional understanding models \cite{ref1, ref2, ref3, ref4, ref5, ref6} or generative models \cite{ref7, ref8, ref9}, unified models seek to simultaneously perform both cross-modal understanding and generation within a single framework \cite{ref10, ref11}. However, current approaches still exhibit significant shortcomings regarding structural consistency and causal generalization.

The real world is inherently multimodal and structured, where diverse modalities—such as text, images, video, and audio—share deep, underlying semantic mechanisms. For instance, the linguistic description "a red cube above a blue sphere" corresponds directly to the spatial relationships between objects in an image and the causal sequence of actions within a video.

Current unified models, however, primarily rely on statistical co-occurrence patterns rather than modeling these intrinsic structural mechanisms \cite{ref4, ref8, ref12, ref13}. This fundamental limitation leads to three critical issues: \textit{(1) Cross-modal Semantic Drift: Inconsistencies when translating or aligning information across different media.} \textit{(2) Weak Compositional Generalization: An inability to accurately synthesize or reason about novel combinations of known concepts.} \textit{(3) Unstable Interventional Generation: A lack of robustness when specific attributes or causal factors are modified.}

Consequently, investigating structure-preserving and causally consistent unified multimodal models is of paramount importance. Developing such frameworks is essential for achieving a higher level of machine intelligence capable of true physical and logical grounding. 

Existing unified models generally fall into two categories:

\textit{Autoregressive-based Unified Models: }The primary advantage of Autoregressive-based Unified Models lies in their inherent flexibility, which naturally supports any-to-any multimodal transitions by treating various data types as sequential tokens \cite{ref15, ref16, ref17}. These models typically adapt from LLMs like LLaMA \cite{ref14, ref15}, Gemma \cite{ref18, ref19, ref20} and Qwen \cite{ref21, ref22, ref23, ref24}. However, this approach faces significant hurdles. Because these models operate primarily on token-level correlations, they often lack explicit structural modeling, which can result in a loss of global coherence \cite{ref25}. Furthermore, they are prone to semantic drift, where cumulative error in sequential prediction leads to outputs that lose their original meaning or logical consistency over long sequences \cite{ref26, ref27, ref28}. 

\textit{Diffusion-based Unified Models: }Diffusion-based Unified Models are highly regarded for their ability to achieve superior generation quality \cite{ref29, ref30, ref31, ref32, ref33}, producing high-fidelity outputs that often surpass autoregressive methods in visual or auditory realism. Nevertheless, they are not without limitations. A major challenge is that their internal structure remains largely uncontrollable, making it difficult to enforce specific spatial or logical constraints \cite{ref34, ref35}. Additionally, these models typically lack reversibility constraints, which, combined with the difficulty of mapping diverse modalities into a unified semantic space, complicates the task of achieving seamless cross-modal integration.

\paragraph{Contributions}

As shown in the Figure 1, we propose C3-UniMM, a unified multimodal modeling framework designed to explicitly incorporate structured causal mechanisms within a single model. This approach addresses critical limitations in existing methods, such as semantic drift, poor compositional generalization, and insufficient cross-modal consistency.

\begin{enumerate}

\item{\bf The Proposed Framework: C3-UniMM}

We propose C3-UniMM, a novel Unified Multimodal Modeling Framework rooted in Structured Causality. The core innovation lies in the explicit introduction of a Structured Latent Causal Graph (SLCG) as a shared semantic intermediary across modalities. By transitioning from traditional direct mapping $x_m \rightarrow x_n$ to a structure-based generative process $x_m \rightarrow G \rightarrow x_n$, the framework explicitly models objects, their relationships, and their underlying generative mechanisms. This approach significantly enhances cross-modal semantic consistency, interpretability, and compositional generalization.

\item{\bf Unified Structural Encoding and Generation}

We design a unified structured encoding and generation mechanism that maps multimodal inputs into a shared causal structural space via slot-based abstraction and structural modeling. This architecture enables the collaborative optimization of multimodal understanding and generation within a single, coherent semantic structure. Furthermore, by incorporating a Unified Decoding Space (UDS), we ensure that generation processes across different modalities share the same semantic manifold, thereby guaranteeing architectural-level consistency and reversibility in cross-modal mapping.

\item{\bf Super Alignment and Causal Consistency}

To further refine the model’s learning objective, we introduce a Super Alignment mechanism and Causal Consistency learning. The Super Alignment operates across three distinct levels—representation, structure, and mechanism—ensuring that the model captures mechanism invariance rather than mere statistical correlations. Coupled with causal cycle consistency and intervention-based training constraints, these mechanisms substantially bolster the model’s robustness and its ability to generalize to out-of-distribution scenarios.

\end{enumerate}

\section{Related Works}

\paragraph{Multimodal understanding model}This category encompasses Vision-Language Transformers \cite{ref4, ref5, ref6}, CLIP-style alignment models \cite{ref26}, and Multimodal Large Language Models (MLLMs) such as LLaVA \cite{ref36, ref37}. These architectures are primarily characterized by their formidable semantic understanding and robust cross-modal alignment capabilities \cite{ref32, ref33, ref38}. However, they typically exhibit weak generative performance when compared to specialized synthesis engines \cite{ref39, ref40}. Furthermore, similar to current AR-based and Diffusion-based systems, these models generally lack explicit structural causal modeling \cite{ref41, ref42, ref43}. This absence of a causal framework limits their ability to reason about the underlying logical relationships between entities, often resulting in outputs that are associative rather than structurally grounded.

\paragraph{Multimodal Generative Model}This category primarily encompasses Text-to-Image Diffusion Models \cite{ref44, ref45, ref46, ref47}, Multimodal Autoregressive (AR) Models \cite{ref48, ref49, ref50, ref51, ref52, ref53, ref54, ref55}, and Unified Token-based Generation Models \cite{ref56, ref57, ref58, ref59, ref60, ref61}. These architectures are distinguished by their formidable generation capabilities, enabling the synthesis of high-fidelity content across diverse modalities. However, they face significant structural limitations. Most notably, they lack mechanistic invariance, meaning they fail to maintain consistent underlying logic across varying contexts or perturbations. Furthermore, these models typically offer no guarantees of cycle consistency, which often results in severe information loss or semantic drift when performing bidirectional transformations (e.g., translating from image to text and back to image).

\paragraph{Causal and Cycle-Consistent Multimodal Frameworks}To bridge the dichotomy between associative understanding and unconstrained generation, emerging research focuses on explicitly modeling the shared generative mechanisms across modalities. In the realm of causal machine learning, foundational works on causal representation learning \cite{ref63} advocate for moving beyond statistical correlations to uncover the invariant mechanisms that govern data generation. In multimodal contexts, early implementations such as counterfactual VQA \cite{ref64} and causal interventions for visual question answering \cite{ref65} have demonstrated the efficacy of structural causal models in mitigating reliance on spurious correlations. Furthermore, ensuring structural stability aligns with the principles of knowledge reduction emphasized in fuzzy rough set theory.

Concurrently, enforcing cycle consistency has proven crucial for maintaining semantic integrity during cross-modal transformations. Pioneered by CycleGAN \cite{ref66} in the vision domain, cycle-consistent adversarial networks established that translating a representation back and forth must retain the original information. This principle has been extended to multimodal tasks by architectures like MirrorGAN \cite{ref67}, which enforces text-to-image-to-text cyclic constraints. Despite these advancements, contemporary unified token-based models (such as Chameleon \cite{ref68} or Show-o \cite{ref69}) still largely rely on token-level fusion and lack explicit structural causal integration. By synthesizing explicit structured latent causal graphs with cycle-consistent bidirectional transformations, our framework aims to guarantee both high-fidelity generation and rigorous semantic stability across any-to-any modal mappings.

\section{Method}

\subsection{Overview}

\begin{figure}[h]
    \centering
    \includegraphics[width=1\textwidth]{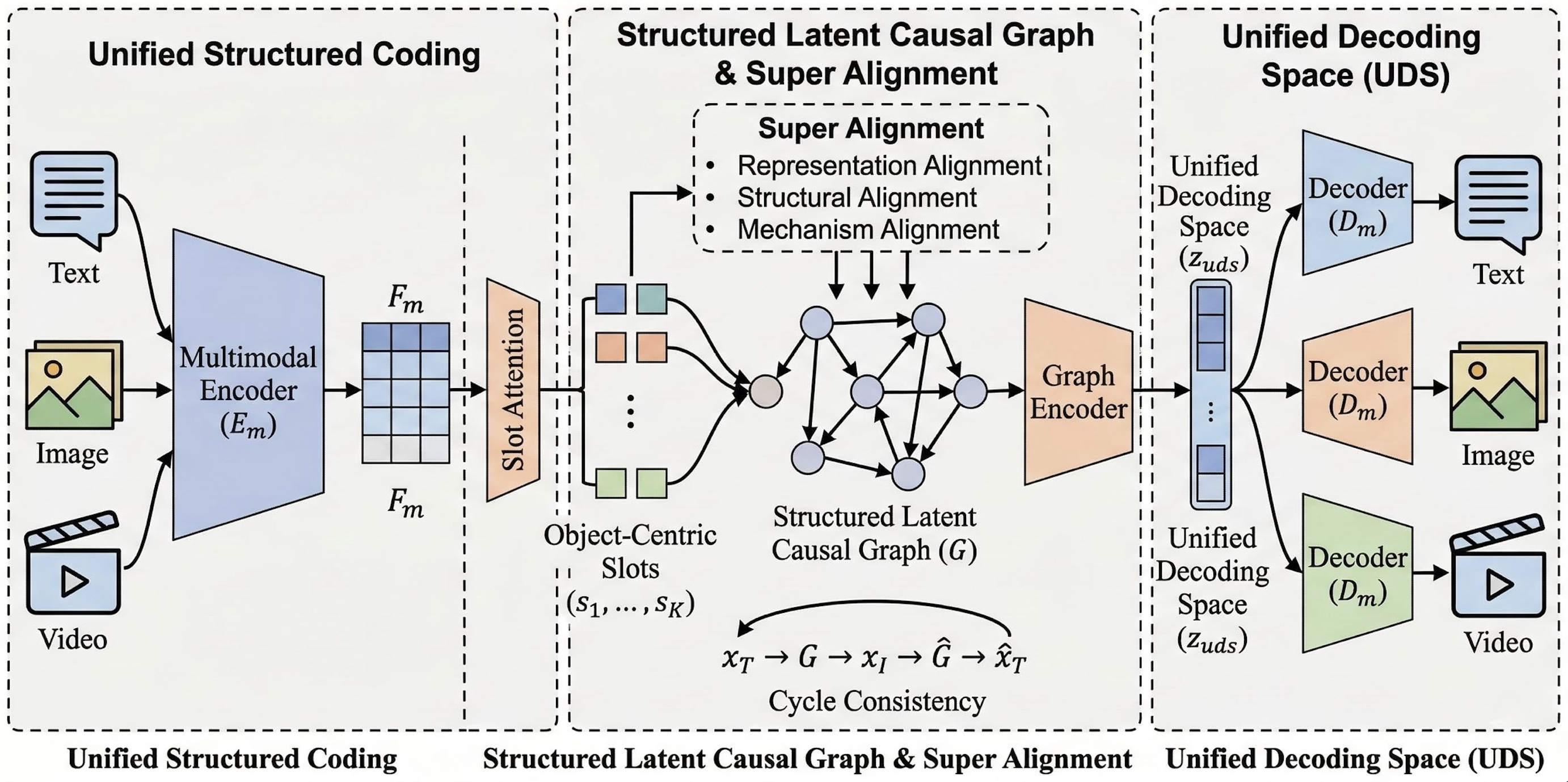}
    \caption{C3-UniMM Architecture. The pipeline consists of three core modules:Unified Structured Coding: Encodes multimodal inputs into Object-Centric Slots ($s_1, \dots, s_K$) using Slot Attention.Structured Latent Causal Graph \& Super Alignment: Maps slots to a Causal Graph ($G$) via Super Alignment (Representation, Structural, and Mechanism) while enforcing Cycle Consistency to minimize semantic drift.Unified Decoding Space (UDS): Projects the graph into a shared latent space ($z_{uds}$) for high-fidelity reconstruction through modality-specific decoders ($D_m$).}
    \label{fig 2}
\end{figure}

Existing unified multimodal models typically learn direct conditional mappings between modalities as follows:\[p_\theta(x_n \mid x_m),\]where $x_m$ denotes the source modality input, $x_n$ denotes the target modality output, and $\theta$ represents the model parameters. This formulation treats cross-modal transformation as a direct mapping from one observation space to another, enabling any-to-any generation and understanding. However, such modeling primarily relies on statistical correlations at the observational level and lacks explicit constraints on the shared generative structure across modalities. Consequently, these models are prone to semantic drift and structural mismatch in tasks involving combinatorial generalization, reversible mapping, and controllable generation.To address these limitations, we propose C3-UniMM as shown in the Figure 2. Our core hypothesis is that observations from different modalities do not correspond to each other directly; instead, they are generated from a single, cross-modally shared latent structure. Let the set of modalities be\[\mathcal{M}=\{T, I, V, A\},\]where $T, I, V,$ and $A$ represent text, image, video, and audio, respectively. For any input sample $x_m \in \mathcal{M}$, we first map it to a shared Structured Latent Variable $G_m$, which then guides the generation of the target modality output $\hat{x}_n$. This process can be unified as:\[G_m = \Phi(E_m(x_m)), \qquad \hat{x}_n = D_n(g(G_m)),\]where $E_m(\cdot)$ is the encoder for modality $m$ to extract input features; $\Phi(\cdot)$ is the Structured Latent Graph Constructor that infers the shared structure from encoded features; $g(\cdot)$ represents the mapping from the latent structural graph to a unified decoding space; $D_n(\cdot)$ is the decoder for the target modality $n$; $G_m$ denotes the latent structural representation induced by input $x_m$; and $\hat{x}_n$ is the generated output.Accordingly, the traditional direct mapping $x_m \rightarrow x_n$ is reformulated into a unified generative process mediated by a shared structure:\[x_m \xrightarrow{E_m, \Phi} G_m \xrightarrow{g, D_n} \hat{x}_n.\]The key to this modeling approach is that the model no longer directly transfers low-level tokens or patch features between modalities. Instead, it extracts composable, intervenable, and cross-modally reusable structural semantics before completing the understanding and generation tasks based on this shared structure. Consequently, C3-UniMM achieves an architectural unification of multimodal representation learning and generative modeling, providing a common semantic interface for subsequent alignment, intervention, and cycle-consistency. The pipeline of C3-UniMM architecture is as shown in Figure 2.

\subsection{Structured Latent Causal Graph and Unified Decoding}

Given an input sample $x_m$ from modality $m$, we first employ a modality-specific encoder to transform the raw observation into a latent feature sequence:
\[
F_m = E_m(x_m) \in \mathbb{R}^{N_m \times d},
\]
where $F_m$ denotes the encoded feature sequence, $N_m$ is the number of tokens, image patches, or spatio-temporal units, and $d$ is the latent dimension. Depending on the input modality, $E_m$ can be instantiated as a Text Transformer, Image ViT, or Video Temporal Transformer. To encourage early-stage cross-modal semantic compatibility, different modality encoders may share high-level projection layers or partially shared parameters.

Although such token-level features preserve rich local information, they often contain modality-specific noise and redundant fine-grained details. Directly aligning or decoding from these features may therefore lead to unstable cross-modal mappings. To obtain a more compact and semantically meaningful representation, we introduce Slot Attention to abstract heterogeneous features into a fixed set of object-level latent units. The resulting Structured Latent Causal Graph (SLCG) is defined as:
\[
G_m=(S_m,A_m), \qquad S_m=\text{Slot}(F_m), \qquad A_m=\sigma(S_m W S_m^\top),
\]
where $S_m=\{s_1,\dots,s_K\}$ denotes $K$ latent slots, with each slot $s_i \in \mathbb{R}^d$ representing an object-level semantic unit. The adjacency matrix $A_m \in \mathbb{R}^{K \times K}$ captures relational dependencies among these slots, $W$ is a learnable bilinear transformation, and $\sigma(\cdot)$ maps relation scores into the $(0,1)$ interval.

This formulation enables the model to represent multimodal inputs not merely as global embeddings, but as structured semantic graphs. In particular, the slot set $S_m$ describes ``what entities are present'', while the adjacency matrix $A_m$ characterizes ``how these entities are related''. Such a decomposition is crucial for preserving object-level structure, supporting compositional reasoning, and reducing semantic drift during cross-modal generation.

After constructing the SLCG, we map the graph representation into a Unified Decoding Space (UDS):
\[
z_m=g(G_m), \qquad \hat{x}_n=D_n(z_m),
\]
where $z_m$ is a shared decoding representation that carries modality-invariant structural semantics, and $D_n(\cdot)$ projects this representation into the observation space of the target modality $n$. Accordingly, the complete generation path from an input modality $m$ to a target modality $n$ can be written as:
\[
\hat{x}_n = D_n(g(\Phi(E_m(x_m)))).
\]

This design reflects the central principle of C3-UniMM: different modalities are generated from a shared structural semantic source rather than through direct observation-to-observation translation. The SLCG and UDS jointly determine the semantic content and relational structure to be generated, while modality-specific decoders determine the final output form. By decoupling shared structural semantics from modality-specific rendering, the proposed framework improves cross-modal consistency, mitigates semantic drift, and enhances the stability and reversibility of multimodal generation.

\subsection{Super Alignment and Causal Consistency Learning}

Shared structural representations alone do not guarantee that different modalities learn consistent internal mechanisms. To address this, we introduce Super Alignment together with causal intervention and cycle-consistency learning, enforcing cross-modal consistency at three levels: representation, structure, and mechanism.

Given latent graphs $G_m=(S_m,A_m)$ and $G_n=(S_n,A_n)$ from modalities $m$ and $n$, we define the cross-modal structural consistency objective as:
\[
\mathcal{C}(m,n)=
\underbrace{\|\bar{G}_m-\bar{G}_n\|_2^2}_{\text{Representation}}
+\alpha\underbrace{\|A_m-A_n\|_F^2}_{\text{Structure}}
+\beta\underbrace{\sum_i\|f_m^i-f_n^i\|_2^2}_{\text{Mechanism}},
\]
where $\bar{G}_m=\text{Pool}(G_m)$ denotes the graph-level representation, $A_m$ and $A_n$ encode relational topologies, and $f_m^i$ represents the generative mechanism of node $i$ under modality $m$. This objective aligns not only semantic outcomes, but also object relations and underlying generative laws.

At the mechanism level, each latent node is modeled under a Structural Causal Model:
\[
s_i=f_i(\text{Pa}(i),\epsilon_i),
\]
where $\text{Pa}(i)$ denotes the parent nodes and $\epsilon_i$ is independent noise. We assume that although observations differ across modalities, the mechanisms governing object states should remain invariant. Thus, Super Alignment encourages mechanism-invariant representations rather than shallow embedding matching.

To improve controllability, we further perform causal intervention on latent slots:
\[
G_m'=do(G_m,s_k=v), \qquad \tilde{x}_n=D_n(g(G_m')),
\]
where editing slot $s_k$ changes the latent structure before decoding it into the target modality. The generated sample is then mapped back to the structural space:
\[
\tilde{G}_n=\Phi(E_n(\tilde{x}_n)),
\]
and ideally satisfies $\tilde{G}_n \approx G_m'$, indicating that structural edits are faithfully reflected in the generated modality.

Finally, to enforce cross-modal reversibility, we construct a cycle path:
\[
\hat{x}_m =
D_m\Big(g\big(\Phi(E_n(D_n(g(\Phi(E_m(x_m))))))\big)\Big).
\]
The model is expected to satisfy:
\[
\hat{x}_m \approx x_m, \qquad
\Phi(E_n(\hat{x}_n)) \approx \Phi(E_m(x_m)),
\]
which correspond to observational cycle reversibility and structural cycle stability, respectively.

Together, Super Alignment, causal intervention, and cycle consistency form the Causal Consistency Learning framework of C3-UniMM. These components jointly enhance semantic consistency, structural controllability, and cross-modal reversibility, enabling stable and interpretable multimodal understanding and generation.

\section{Experiments}

\subsection{Benchmarks and Implementation Details}

C3-UniMM is evaluated across image/video understanding, image/video generation, and compositional generalization using a broad set of benchmarks. We adopt MSCOCO for semantic alignment, Visual Genome for relational modeling, and CLEVR for structural consistency and causal reasoning. For generation, we use GenEval and DPG-Bench to assess structured image synthesis, VBench for video temporal stability, and VideoMME for long-form video understanding. Zero-shot compositional tests are further included to evaluate generalization under novel object-attribute and relational combinations.

For implementation, C3-UniMM employs ViT and Transformer backbones to encode visual and textual inputs into a unified latent space. Slot Attention abstracts the features into 16 object-level nodes, which form the Structured Latent Causal Graph (SLCG). The graph is then mapped into a Unified Decoding Space (UDS) with a hidden dimension of 1024 for cross-modal generation. The model is trained for 300K steps with AdamW and mixed precision, jointly optimizing generation, Super Alignment, causal cycle-consistency, and intervention consistency losses.

We evaluate the model from three perspectives: understanding, generation, and structural preservation. Accuracy and CIDEr are used for vision-language understanding and text quality, while FID and CLIPScore measure image realism and semantic alignment. GenEval and DPG-Bench assess attribute binding and relational control, and VBench evaluates temporal coherence. We additionally report Compositional Accuracy and Semantic Drift Score (SDS) to quantify zero-shot generalization and structural consistency during cross-modal transformations.

\subsection{Experimental Results and Analysis}

We evaluate C3-UniMM against state-of-the-art methods across several representative multi-modal understanding benchmarks, including MME, GQA, SEED, MMB, MMMU, MMStar, and AI2D.

In the following tables, \#Params denotes the parameter scale of the base model. "Und. Only" refers to models designed exclusively for understanding tasks, while "Native Unified" represents unified multi-modal models that natively support both understanding and generation. Unless otherwise specified, all results are reported under the zero-shot setting.

\begin{table}[H]
\centering
\caption{Benchmark results comparing C3-UniMM against various multimodal models across understanding and unified benchmarks.}
\label{tab:multimodal-benchmarks}
\resizebox{\textwidth}{!}{
\begin{tabular}{llcccccccc}
\toprule
\textbf{Types} & \textbf{Models} & \textbf{\#Params} & \textbf{MME$\uparrow$} & \textbf{GQA$\uparrow$} & \textbf{SEED$\uparrow$} & \textbf{MMB$\uparrow$} & \textbf{MMMU$\uparrow$} & \textbf{MMStar$\uparrow$} & \textbf{AI2D$\uparrow$} \\ 
\midrule
\textbf{Und. Only} & LLaVA-v1.5 \cite{ref70} & 7B & 1510.7 & 62 & 58.6 & 64.3 & - & - & - \\
 & Qwen-VL-Chat \cite{ref71} & 7B & 1487.6 & 57.5 & 58.2 & 60.6 & - & - & 57.7 \\
 & LLaVA-OV \cite{ref72} & 7B & 1580 & - & - & 80.8 & 48.8 & 57.5 & 81.4 \\
\midrule
\textbf{Unify via} & NExT-GPT \cite{ref73} & 13B & - & - & 57.5 & 58 & - & - & - \\
\textbf{Assembling} & SEED-X \cite{ref74} & 17B & 1457 & 49.1 & 66.5 & 70.1 & 35.6 & - & - \\
 & MetaMorph \cite{ref75} & 8B & - & - & 71.8 & 75.2 & - & - & - \\
 & TokenFlow-XL \cite{ref76} & 14B & 1551.1 & 62.5 & 72.6 & 76.8 & 43.2 & - & 75.9 \\
 & ILLUME \cite{ref77} & 7B & 1445.3 & - & 72.9 & 75.1 & 38.2 & - & 71.4 \\
\midrule
\textbf{Native Unified} & BAGEL \cite{ref78} & 14B & 1687 & - & - & 85 & 55.3 & - & - \\
 & Show-o \cite{ref79} & 1.3B & 1097.2 & 58 & 51.5 & - & 27.4 & - & - \\
 & JanusFlow \cite{ref80} & 1.5B & 1333.1 & 60.3 & 70.5 & 74.9 & 29.3 & - & - \\
 & SynerGen-VL \cite{ref81} & 2.4B & 1381 & - & - & 53.7 & 34.2 & - & - \\
 & Janus-Pro \cite{ref82} & 1.5B & 1444 & 59.3 & 68.3 & 75.5 & 36.3 & - & - \\
 & Show-o2 \cite{ref83} & 1.5B & 1450.9 & 60 & 65.6 & 67.4 & 37.1 & 43.4 & 69 \\
 & Emu3 \cite{ref84} & 8B & - & 60.3 & 68.2 & 58.5 & 31.6 & - & 70 \\
 & VILA-U \cite{ref85} & 7B & 1401.8 & 60.8 & 59 & - & - & - & - \\
 & MUSE-VL \cite{ref86} & 7B & - & - & 69.1 & 72.1 & 39.7 & 49.6 & 69.8 \\
 & Liquid \cite{ref87} & 8B & 1448 & 61.1 & - & - & - & - & - \\
 & Janus-Pro \cite{ref82} & 7B & 1567.1 & 62 & 72.1 & 79.2 & 41 & - & - \\
 & Moga \cite{ref130} & 7B & 1592 & 60.9 & 74.6 & 75 & 44.2 & - & - \\
 & Show-o2 \cite{ref83} & 7B & 1620.5 & 63.1 & 69.8 & 79.3 & 48.9 & 56.6 & 78.6 \\
\cmidrule{2-10}
 & \textbf{C3-UniMM (Ours)} & \textbf{1.5B} & \textbf{1485.2} & \textbf{61.2} & \textbf{69.4} & \textbf{72.8} & \textbf{42.6} & \textbf{47.9} & \textbf{71.5} \\
 & \textbf{C3-UniMM (Ours)} & \textbf{7B} & \textbf{1678.3} & \textbf{64.5} & \textbf{74.1} & \textbf{83.6} & \textbf{54.7} & \textbf{61.8} & \textbf{81.2} \\
\bottomrule
\end{tabular}
}
\end{table}

As shown in Table 1, C3-UniMM establishes a new state-of-the-art for native unified models. Most notably, the 7B variant outperforms significantly larger models (like the 14B BAGEL and 17B SEED-X) across nearly every metric, including MME, MMMU, and MMB. Even the 1.5B version remains highly competitive, often matching or exceeding the performance of 7B and 8B models from other families, demonstrating superior architectural efficiency in multimodal understanding.

\begin{table}[H]
\centering
\caption{Performance comparison of our model against existing models on various video benchmarks.}
\label{tab:video-benchmarks}
\resizebox{\textwidth}{!}{
\begin{tabular}{llccccccc}
\toprule
\textbf{Model} & \textbf{\#Params} & \textbf{\#Frames} & \textbf{ActNet-QA} & \textbf{MVBench} & \textbf{NExT-QA} & \textbf{Perception} & \textbf{LongVideoBench} & \textbf{VideoMME} \\ 
\midrule
GPT-4V \cite{ref93} & - & - & 57 & 43.5 & - & - & 61.3 & 59.9/63.3 \\
GPT-4o \cite{ref94} & - & - & - & - & - & - & 66.7 & 71.9/77.2 \\
Gemini-1.5-Flash \cite{ref95} & - & - & 55.3 & - & - & - & 61.6 & 70.3/75.0 \\
Gemini-1.5-Pro \cite{ref95} & - & - & 57.5 & - & - & - & 64 & 75.0/81.3 \\
VILA \cite{ref96} & 40B & - & 58 & - & 67.9 & 54 & - & 60.1/61.1 \\
PLLaVA \cite{ref97} & 34B & 16 & 60.9 & 58.1 & - & - & 53.2 & - \\
LongVA \cite{ref98} & 7B & - & 50 & - & 68.3 & - & - & 52.6/54.3 \\
IXC-2.5 \cite{ref99} & 7B & 64 & 52.8 & 69.1 & 71 & 34.4 & - & 55.8/58.8 \\
LLaVA-OV \cite{ref72} & 7B & 32 & 56.6 & 56.7 & 79.4 & 57.1 & 56.5 & 58.2/61.5 \\
VideoLLaMA2 \cite{ref100} & 7B & 16 & 50.2 & 54.6 & - & 51.4 & - & 47.9/50.3 \\
Show-o2$^\dagger$ \cite{ref83} & 1.5B & 32 & 52.7 & 49.8 & 72.1 & 56.1 & 49.2 & 48.0/51.6 \\
Show-o2$^\dagger$ \cite{ref83} & 7B & 16/32 & 56.4 & 55.8 & 79 & 61.9 & 55.5 & 57.4/60.9 \\ 
\midrule
\textbf{C3-UniMM (Ours)} & \textbf{1.5B} & \textbf{32} & \textbf{55.6} & \textbf{52.3} & \textbf{75.4} & \textbf{58.7} & \textbf{52.6} & \textbf{52.1/55.8} \\
\textbf{C3-UniMM (Ours)} & \textbf{7B} & \textbf{16/32} & \textbf{59.8} & \textbf{59.1} & \textbf{81.7} & \textbf{64.8} & \textbf{58.9} & \textbf{60.5/64.2} \\
\bottomrule
\end{tabular}
}
\end{table}

As shown in Table 2, the results indicate that C3-UniMM is a top-tier contender in video comprehension. The 7B model achieves a VideoMME score of 60.5/64.2, surpassing established models like VILA (40B) and LLaVA-OV (7B). Its high performance on NExT-QA (81.7) and Perception (64.8) suggests a robust ability to handle temporal dynamics and complex visual reasoning in video sequences.

\begin{table}[H]
\centering
\caption{Benchmark results comparing our C3-UniMM model against existing Gen-Only, Assembling, and Native Unified methods across various visual attributes.}
\label{tab:benchmark-results-2}
\resizebox{\textwidth}{!}{
\begin{tabular}{llccccccccc}
\toprule
\textbf{Type} & \textbf{Method} & \textbf{\#Params} & \textbf{\#Data} & \textbf{Single} & \textbf{Two} & \textbf{Count} & \textbf{Colors} & \textbf{Position} & \textbf{Color Attr} & \textbf{Overall$\uparrow$} \\ 
\midrule
Gen Only & SD3-Medium \cite{ref131} & - & - & 0.99 & 0.94 & 0.72 & 0.89 & 0.33 & 0.6 & 0.74 \\ 
\midrule
\multirow{4}{*}{Assembling} & SEED-X \cite{ref74} & 17B & 158M+ & 0.97 & 0.58 & 0.26 & 0.8 & 0.19 & 0.14 & 0.49 \\
 & TokenFlow-XL \cite{ref76} & 14B & 60M+ & 0.95 & 0.6 & 0.41 & 0.81 & 0.16 & 0.24 & 0.55 \\
 & ILLUME \cite{ref77} & 7B & 15M+ & 0.99 & 0.86 & 0.45 & 0.71 & 0.39 & 0.28 & 0.61 \\
 & MetaQuery-XL \cite{ref90} & 7B & 28M+ & - & - & - & - & - & - & 0.8 \\ 
\midrule
\multirow{12}{*}{Native Unified} & Show-o \cite{ref79} & 1.3B & 2.0B & 0.98 & 0.8 & 0.66 & 0.84 & 0.31 & 0.5 & 0.68 \\
 & Emu3 \cite{ref84} & 8B & - & - & - & - & - & - & - & 0.66 \\
 & MUSE-VL \cite{ref86} & 7B & 24M & - & - & - & - & - & - & 0.57 \\
 & Transfusion \cite{ref91} & 7B & 3.5B & - & - & - & - & - & - & 0.63 \\
 & D-DiT \cite{ref92} & 2B & 40M & 0.97 & 0.8 & 0.54 & 0.76 & 0.32 & 0.5 & 0.65 \\
 & Janus-Pro \cite{ref82} & 7B & 144M & 0.99 & 0.89 & 0.59 & 0.9 & 0.79 & 0.66 & 0.8 \\
 & BAGEL \cite{ref78} & 14B & 1600M & 0.98 & 0.95 & 0.84 & 0.95 & 0.78 & 0.77 & 0.88 \\
 & Moga \cite{ref130} & 7B & - & 1 & 0.97 & 0.83 & 0.93 & 0.84 & 0.8 & 0.89 \\
 & Show-o2 \cite{ref83} & 1.5B & 66M & 0.99 & 0.86 & 0.55 & 0.86 & 0.46 & 0.63 & 0.73 \\
 & Show-o2 \cite{ref83} & 7B & 66M & 1 & 0.87 & 0.58 & 0.92 & 0.52 & 0.62 & 0.76 \\ \cmidrule{2-11}
 & \textbf{C3-UniMM (Ours)} & \textbf{1.5B} & \textbf{66M} & \textbf{1} & \textbf{0.9} & \textbf{0.62} & \textbf{0.88} & \textbf{0.53} & \textbf{0.68} & \textbf{0.78} \\
 & \textbf{C3-UniMM (Ours)} & \textbf{7B} & \textbf{66M} & \textbf{1} & \textbf{0.93} & \textbf{0.66} & \textbf{0.94} & \textbf{0.6} & \textbf{0.72} & \textbf{0.82} \\
\bottomrule
\end{tabular}
}
\end{table}

As shown in Table 3, C3-UniMM demonstrates exceptional precision in text-to-image generation. It achieves an Overall score of 0.82, outperforming specialized "Gen-Only" models like SD3-Medium and other unified models such as Show-o2 and Janus-Pro. The model is particularly strong at handling Single and Two-object compositions, where it achieves near-perfect or perfect scores (1.0).

\begin{table}[H]
\centering
\caption{Comprehensive performance comparison of various models across multiple evaluation metrics.}
\label{tab:comprehensive-metrics}
\resizebox{\textwidth}{!}{
\begin{tabular}{llcccccccccccc}
\toprule
\textbf{Models} & \textbf{\#Params} & \textbf{QS} & \textbf{SS} & \textbf{SC} & \textbf{BC} & \textbf{TF} & \textbf{MS} & \textbf{DD} & \textbf{AQ} & \textbf{IQ} & \textbf{OC} & \textbf{MO} & \textbf{HA} \\ 
\midrule
ModelScope \cite{ref113} & 1.7B & 75.75 & 78.05 & 66.54 & 89.87 & 95.29 & 98.28 & 95.79 & 66.39 & 52.06 & 58.57 & 82.25 & 38.98 \\
LaVie \cite{ref114} & 3B & 77.08 & 78.78 & 70.31 & 91.41 & 97.47 & 98.30 & 96.38 & 49.72 & 54.94 & 61.90 & 91.82 & 33.32 \\
OpenSoraPlan \cite{ref115} & - & 77.23 & 80.14 & 65.62 & 97.79 & 97.24 & 99.20 & 99.05 & 30.28 & 60.42 & 56.21 & 85.56 & 43.58 \\
Show-1 \cite{ref116} & 6B & 78.93 & 80.42 & 72.98 & 95.53 & 98.02 & 99.12 & 98.24 & 44.44 & 57.35 & 58.66 & 93.07 & 45.47 \\
AnimateDiff \cite{ref117} & - & 80.27 & 82.90 & 69.75 & 95.30 & 97.68 & 98.75 & 97.76 & 40.83 & 67.16 & 70.10 & 90.90 & 36.88 \\
VideoCrafter2 \cite{ref118} & - & 80.44 & 82.20 & 73.42 & 96.85 & 98.22 & 98.41 & 97.73 & 42.50 & 63.13 & 67.22 & 92.55 & 40.66 \\
CogVideoX \cite{ref119} & 5B & 81.61 & 82.75 & 77.04 & 96.23 & 96.52 & 98.66 & 96.92 & 70.97 & 61.98 & 62.90 & 85.23 & 62.11 \\
Step-Video-T2V \cite{ref120} & 30B & 81.83 & 84.46 & 71.28 & 98.06 & 97.67 & 99.40 & 99.08 & 53.06 & 61.23 & 70.63 & 80.56 & 50.55 \\
Gen-3 \cite{ref121} & - & 82.32 & 84.11 & 75.17 & 97.10 & 96.62 & 98.61 & 99.23 & 60.14 & 63.34 & 66.82 & 87.81 & 53.64 \\
Emu3 \cite{ref84} & 8B & 80.96 & - & - & 95.32 & 97.69 & - & 98.93 & 79.27 & 59.64 & - & 86.17 & 44.64 \\
VILA-U \cite{ref85} & 7B & 74.01 & 76.26 & 65.04 & - & - & - & - & - & - & - & - & - \\
HAPLO \cite{ref85} & 9B & 78.10 & - & - & 96.40 & 97.60 & - & 96.80 & 65.30 & - & - & - & - \\
Show-o2 \cite{ref83} & 2B & 81.34 & 82.10 & 78.31 & 97.28 & 96.78 & 97.68 & 98.25 & 40.83 & 65.15 & 67.06 & 94.81 & 76.01 \\
\textbf{C3-UniMM (Ours)} & \textbf{2B} & \textbf{83.12} & \textbf{84.35} & \textbf{81.74} & \textbf{98.06} & \textbf{97.11} & \textbf{98.42} & \textbf{98.96} & \textbf{44.72} & \textbf{67.83} & \textbf{69.55} & \textbf{95.64} & \textbf{78.90} \\
\bottomrule
\end{tabular}
}
\end{table}

As shown in Table 4, Across a wide battery of video metrics, C3-UniMM (2B) consistently ranks at the top, particularly in Quality Score (QS: 83.12) and Motion Smoothness (MO: 95.64). Its high Human Alignment (HA: 78.90) score suggests that its generated videos are not just technically sound but also more satisfying to human observers compared to models like CogVideoX or Gen-3.

\begin{table}[H]
\centering
\caption{Performance evaluation of various models across different image-to-video (I2V) metrics.}
\label{tab:i2v-metrics}
\resizebox{\textwidth}{!}{
\begin{tabular}{lcccccccccc}
\toprule
\textbf{Models} & \textbf{I2V Subject} & \textbf{I2V Background} & \textbf{Camera Motion} & \textbf{Subject Consistency} & \textbf{Background Consistency} & \textbf{Temporal Flickering} & \textbf{Motion Smoothness} & \textbf{Dynamic Degree} & \textbf{Aesthetic Quality} & \textbf{Imaging Quality} \\ 
\midrule
DynamiCrafter \cite{ref122} & 96.71 & 96.05 & 35.44 & 95.69 & 97.38 & 97.63 & 97.38 & 47.40 & 66.46 & 69.34 \\
SEINE \cite{ref123} & 94.85 & 94.02 & 23.36 & 94.20 & 97.26 & 96.72 & 96.68 & 34.31 & 58.42 & 70.97 \\
I2VGen-XL \cite{ref124} & 96.74 & 95.44 & 13.32 & 96.36 & 97.93 & 98.48 & 98.31 & 24.96 & 65.33 & 69.85 \\
Animate-Anything \cite{ref125} & 98.54 & 96.88 & 12.56 & 98.90 & 98.19 & 98.14 & 98.61 & 2.68 & 67.12 & 72.09 \\
ConsistI2V \cite{ref126} & 94.69 & 94.57 & 33.60 & 95.27 & 98.28 & 97.56 & 97.38 & 18.62 & 59.00 & 66.92 \\
VideoCrafter-I2V \cite{ref127} & 90.97 & 90.51 & 33.58 & 97.86 & 98.79 & 98.19 & 98.00 & 22.60 & 60.78 & 71.68 \\
SVD-XT \cite{ref128} & 97.51 & 97.62 & - & 95.42 & 96.77 & 99.17 & 98.12 & 43.17 & 60.23 & 70.23 \\
MarDini \cite{ref129} & 98.78 & 96.46 & - & - & - & - & - & - & - & - \\
Show-o2 \cite{ref83} & 96.94 & 98.83 & 28.41 & 93.83 & 97.45 & - & 97.76 & 25.85 & 61.92 & 69.87 \\
\textbf{C3-UniMM (Ours)} & \textbf{97.88} & \textbf{99.12} & \textbf{30.96} & \textbf{95.94} & \textbf{98.26} & \textbf{98.21} & \textbf{98.34} & \textbf{29.47} & \textbf{64.73} & \textbf{72.11} \\
\bottomrule
\end{tabular}
}
\end{table}

As shown in Table 5, the model excels in consistency and quality. It reaches a Background Consistency of 99.12 and Imaging Quality of 72.11, outperforming specialized I2V models like SVD-XT and DynamicCrafter. This suggests that the model is exceptionally good at preserving the identity and environmental details of a source image while animating it.

\begin{table}[H]
\centering
\caption{Ablation Study of C3-UniMM Model Components}
\label{tab:ablation_study}
\begin{tabular}{lcccc}
\toprule
\textbf{Model Variant} & \textbf{MMMU$\uparrow$} & \textbf{SEED$\uparrow$} & \textbf{\begin{tabular}[c]{@{}c@{}}GenEval\\ (Overall)$\uparrow$\end{tabular}} & \textbf{\begin{tabular}[c]{@{}c@{}}DPG-Bench\\ (Overall)$\uparrow$\end{tabular}} \\ \midrule
\textbf{C3-UniMM (Full)} & \textbf{54.7} & \textbf{74.1} & \textbf{0.82} & \textbf{89.84} \\
w/o Graph                & 49.1          & 70.2          & 0.76          & 86.21          \\
w/o Super Alignment      & 50.3          & 71            & 0.77          & 86.95          \\
w/o Cycle Loss           & 52.6          & 72.4          & 0.8           & 88.07          \\
w/o UDS                  & 51.8          & 71.9          & 0.79          & 87.42          \\ \bottomrule
\end{tabular}
\end{table}

The ablation trials presented in Table 6 demonstrate that the full C3-UniMM configuration delivers peak performance, as removing any individual component results in a measurable decline across all benchmarks. The Graph component is identified as the most critical element for structural understanding, showing the steepest performance drops upon removal, while Super Alignment and UDS remain essential for balancing the model's generation and comprehension capabilities. Although Cycle Loss has a comparatively smaller impact than the Graph component, it remains a necessary contributor to the model's overall stability and scoring consistency.

\section{Conclusion}
The C3-UniMM framework marks a significant departure from standard multimodal AI by moving beyond statistical correlations toward a foundational Structured Latent Causal Graph (SLCG). While traditional models often struggle with logical inconsistencies, C3-UniMM anchors its reasoning in the physical laws and common sense of the real world. By utilizing slot-based abstraction, the architecture projects disparate data streams into a unified, shared causal latent space. This shared environment allows the model to perform both understanding and generative tasks synergistically, ensuring that every output is logically derived from the same causal structures used to interpret the input.

To maintain the reliability of these mappings, the framework employs a Super Alignment mechanism designed to enforce causal integrity. Through causal cycle consistency and intervention-based constraints, the model ensures its internal logic remains invariant even when specific variables shift. This robustness is critical for the framework’s transition into Embodied Intelligence. By integrating Automated Causal Discovery, future iterations will enable unsupervised structure extraction, scaling the model to incorporate 3D point clouds and tactile signals. Ultimately, this research aims for temporal stability, allowing AI agents to navigate and interact safely within dynamic, contact-rich physical environments.
\appendix

\bibliographystyle{unsrtnat}
\bibliography{main}

@inproceedings{ref1,
  author = {Liang, Z. and Xu, Y. and Hong, Y. and Shang, P. and Wang, Q. and Fu, Q. and Liu, K.},
  title = {A survey of multimodel large language models},
  booktitle = {Proceedings of the 3rd International Conference on Computer, Artificial Intelligence and Control Engineering},
  pages = {405--409},
  year = {2024}
}

@article{ref2,
  author = {Yin, S. and Fu, C. and Zhao, S. and Li, K. and Sun, X. and Xu, T. and Chen, E.},
  title = {A survey on multimodal large language models},
  journal = {National Science Review},
  volume = {11},
  number = {12},
  pages = {nwae403},
  year = {2024}
}

@article{ref3,
  author = {Carolan, K. and Fennelly, L. and Smeaton, A. F.},
  title = {A review of multi-modal large language and vision models},
  journal = {arXiv preprint arXiv:2404.01322},
  year = {2024}
}

@article{ref4,
  author = {Bordes, F. and Pang, R. Y. and Ajay, A. and Li, A. C. and Bardes, A. and Petryk, S. and Ma{\~n}as, O. and Lin, Z. and Mahmoud, A. and Jayaraman, B. and others},
  title = {An introduction to vision-language modeling},
  journal = {arXiv preprint arXiv:2405.17247},
  year = {2024}
}

@article{ref5,
  author = {Hartsock, I. and Rasool, G.},
  title = {Vision-language models for medical report generation and visual question answering: A review},
  journal = {Frontiers in Artificial Intelligence},
  volume = {7},
  pages = {1430984},
  year = {2024}
}

@article{ref6,
  author = {Li, Z. and Wu, X. and Du, H. and Liu, F. and Nghiem, H. and Shi, G.},
  title = {A survey of state of the art large vision language models: Alignment, benchmark, evaluations and challenges},
  journal = {Preprint},
  year = {2024}
}

@inproceedings{ref7,
  author = {Ho, J. and Jain, A. and Abbeel, P.},
  title = {Denoising diffusion probabilistic models},
  booktitle = {Advances in neural information processing systems},
  volume = {33},
  pages = {6840--6851},
  year = {2020}
}

@inproceedings{ref8,
  author = {Sohl-Dickstein, J. and Weiss, E. and Maheswaranathan, N. and Ganguli, S.},
  title = {Deep unsupervised learning using nonequilibrium thermodynamics},
  booktitle = {International conference on machine learning},
  publisher = {PMLR},
  pages = {2256--2265},
  year = {2015}
}

@article{ref9,
  author = {Cao, P. and Zhou, F. and Song, Q. and Yang, L.},
  title = {Controllable generation with text-to-image diffusion models: A survey},
  journal = {arXiv preprint arXiv:2403.04279},
  year = {2024}
}

@article{ref10,
  author = {Li, S. and Kallidromitis, K. and Gokul, A. and Liao, Z. and Kato, Y. and Kozuka, K. and Grover, A.},
  title = {Omniflow: Any-to-any generation with multi-modal rectified flows},
  journal = {arXiv preprint arXiv:2412.01169},
  year = {2024}
}

@article{ref11,
  author = {Wang, P. and Peng, Y. and Gan, Y. and Hu, L. and Xie, T. and Wang, X. and Wei, Y. and Tang, C. and Zhu, B. and Li, C. and others},
  title = {Skywork unipic: Unified autoregressive modeling for visual understanding and generation},
  journal = {arXiv preprint arXiv:2508.03320},
  year = {2025}
}

@article{ref12,
  author = {Chen, J. and Xu, Z. and Pan, X. and Hu, Y. and Qin, C. and Goldstein, T. and Huang, L. and Zhou, T. and Xie, S. and Savarese, S. and others},
  title = {Blip3-o: A family of fully open unified multimodal models-architecture, training and dataset},
  journal = {arXiv preprint arXiv:2505.09568},
  year = {2025}
}

@article{ref13,
  author = {Wu, S. and Wu, Z. and Gong, Z. and Tao, Q. and Jin, S. and Li, Q. and Li, W. and Loy, C. C.},
  title = {Openuni: A simple baseline for unified multimodal understanding and generation},
  journal = {arXiv preprint arXiv:2505.23661},
  year = {2025}
}

@article{ref14,
  author = {Touvron, H. and Lavril, T. and Izacard, G. and Martinet, X. and Lachaux, M.-A. and Lacroix, T. and Rozi{\`e}re, B. and Goyal, N. and Hambro, E. and Azhar, F. and others},
  title = {Llama: Open and efficient foundation language models},
  journal = {arXiv preprint arXiv:2302.13971},
  year = {2023}
}

@article{ref15,
  author = {Touvron, H. and Martin, L. and Stone, K. and Albert, P. and Almahairi, A. and Babaei, Y. and Bashlykov, N. and Batra, S. and Bhargava, P. and Bhosale, S. and others},
  title = {Llama 2: Open foundation and fine-tuned chat models},
  journal = {arXiv preprint arXiv:2307.09288},
  year = {2023}
}

@article{ref16,
  author = {Tang, H. and Liu, H. and Xiao, X.},
  title = {Ugen: Unified autoregressive multimodal model with progressive vocabulary learning},
  journal = {arXiv preprint arXiv:2503.21193},
  year = {2025}
}

@article{ref17,
  author = {Chern, E. and Su, J. and Ma, Y. and Liu, P.},
  title = {Anole: An open, autoregressive, native large multimodal models for interleaved image-text generation},
  journal = {arXiv preprint arXiv:2407.06135},
  year = {2024}
}

@article{ref18,
  author = {{Gemma Team} and Mesnard, T. and Hardin, C. and Dadashi, R. and Bhupatiraju, S. and Pathak, S. and Sifre, L. and Rivi{\`e}re, M. and Kale, M. S. and Love, J. and others},
  title = {Gemma: Open models based on gemini research and technology},
  journal = {arXiv preprint arXiv:2403.08295},
  year = {2024}
}

@article{ref19,
  author = {{Gemma Team} and Riviere, M. and Pathak, S. and Sessa, P. G. and Hardin, C. and Bhupatiraju, S. and Hussenot, L. and Mesnard, T. and Shahriari, B. and Ram{\'e}, A. and others},
  title = {Gemma 2: Improving open language models at a practical size},
  journal = {arXiv preprint arXiv:2408.00118},
  year = {2024}
}

@article{ref20,
  author = {{Gemma Team} and Kamath, A. and Ferret, J. and Pathak, S. and Vieillard, N. and Merhej, R. and Perrin, S. and Matejovicova, T. and Ram{\'e}, A. and Rivi{\`e}re, M. and others},
  title = {Gemma 3 technical report},
  journal = {arXiv preprint arXiv:2503.19786},
  year = {2025}
}

@article{ref21,
  author = {Bai, J. and Bai, S. and Yang, S. and Wang, S. and Tan, S. and Wang, P. and Lin, J. and Zhou, C. and Zhou, J.},
  title = {Qwen-vl: A versatile vision-language model for understanding, localization, text reading, and beyond},
  journal = {arXiv preprint arXiv:2409.12191},
  year = {2024}
}

@article{ref22,
  author = {Yang, A. and Yang, B. and Zhang, B. and Hui, B. and Zheng, B. and Yu, B. and Li, C. and Liu, D. and Huang, F. and Wei, H. and others},
  title = {Qwen2.5 technical report},
  journal = {arXiv preprint arXiv:2412.15115},
  year = {2024}
}

@article{ref23,
  author = {Wang, P. and Bai, S. and Tan, S. and Wang, S. and Fan, Z. and Bai, J. and Chen, K. and Liu, X. and Wang, J. and Ge, W. and others},
  title = {Qwen2-vl: Enhancing vision-language model’s perception of the world at any resolution},
  journal = {arXiv preprint arXiv:2409.12191},
  year = {2024}
}

@article{ref24,
  author = {Bai, S. and Chen, K. and Liu, X. and Wang, J. and Ge, W. and Song, S. and Dang, K. and Wang, P. and Wang, S. and Tang, J. and others},
  title = {Qwen2.5-vl technical report},
  journal = {arXiv preprint arXiv:2502.13923},
  year = {2025}
}

@article{ref25,
  author = {Jiao, Y. and Qiu, H. and Jie, Z. and Chen, S. and Chen, J. and Ma, L. and Jiang, Y.G.},
  title = {Unitoken: Harmonizing multimodal understanding and generation through unified visual encoding},
  journal = {arXiv preprint arXiv:2504.04423},
  year = {2025}
}

@inproceedings{ref26,
  author = {Radford, A. and Kim, J. W. and Hallacy, C. and Ramesh, A. and Goh, G. and Agarwal, S. and Sastry, G. and Askell, A. and Mishkin, P. and Clark, J. and others},
  title = {Learning transferable visual models from natural language supervision},
  booktitle = {International conference on machine learning},
  publisher = {PMLR},
  pages = {8748--8763},
  year = {2021}
}

@inproceedings{ref27,
  author = {Zhai, X. and Mustafa, B. and Kolesnikov, A. and Beyer, L.},
  title = {Sigmoid loss for language image pre-training},
  booktitle = {Proceedings of the IEEE/CVF international conference on computer vision},
  pages = {11975--11986},
  year = {2023}
}

@article{ref28,
  author = {Sun, Q. and Fang, Y. and Wu, L. and Wang, X. and Cao, Y.},
  title = {Eva-clip: Improved training techniques for clip at scale},
  journal = {arXiv preprint arXiv:2303.15389},
  year = {2023}
}

@article{ref29,
  author = {Dhariwal, P. and Nichol, A.},
  title = {Diffusion models beat gans on image synthesis},
  journal = {Advances in neural information processing systems},
  volume = {34},
  pages = {8780--8794},
  year = {2021}
}

@inproceedings{ref30,
  author = {Ho, J. and Salimans, T.},
  title = {Classifier-free diffusion guidance},
  booktitle = {NeurIPS 2021 Workshop on Deep Generative Models and Downstream Applications},
  year = {2021}
}

@inproceedings{ref31,
  author = {Nichol, A. Q. and Dhariwal, P.},
  title = {Improved denoising diffusion probabilistic models},
  booktitle = {International conference on machine learning},
  publisher = {PMLR},
  pages = {8162--8171},
  year = {2021}
}

@inproceedings{ref32,
  author = {Song, J. and Meng, C. and Ermon, S.},
  title = {Denoising diffusion implicit models},
  booktitle = {International Conference on Learning Representations},
  year = {2021}
}

@article{ref33,
  author = {Lu, C. and Zhou, Y. and Bao, F. and Chen, J. and Li, C. and Zhu, J.},
  title = {Dpm-solver: A fast ode solver for diffusion probabilistic model sampling in around 10 steps},
  journal = {Advances in Neural Information Processing Systems},
  volume = {35},
  pages = {5775--5787},
  year = {2022}
}

@article{ref34,
  author = {{I. Labs} and Khanna, S. and Kharbanda, S. and Li, S. and Varma, H. and Wang, E. and Birnbaum, S. and Luo, Z. and Miraoui, Y. and Palrecha, A. and Ermon, S. and Grover, A. and Kuleshov, V.},
  title = {Mercury: Ultra-fast language models based on diffusion},
  journal = {arXiv preprint arXiv:2506.17298},
  year = {2025}
}

@misc{ref35,
  author = {{Google Deepmind}},
  title = {Gemini diffusion},
  howpublished = {\url{https://deepmind.google/models/gemini-diffusion}},
  year = {2025},
  note = {Accessed: 2025-06-25}
}

@inproceedings{ref36,
  author = {Liu, H. and Li, C. and Li, Y. and Lee, Y. J.},
  title = {Improved baselines with visual instruction tuning},
  booktitle = {Proceedings of the IEEE/CVF Conference on Computer Vision and Pattern Recognition},
  pages = {26296--26306},
  year = {2024}
}

@misc{ref37,
  author = {Liu, H. and Li, C. and Li, Y. and Li, B. and Zhang, Y. and Shen, S. and Lee, Y. J.},
  title = {Llavanext: Improved reasoning, ocr, and world knowledge},
  year = {2024}
}

@inproceedings{ref38,
  author = {Gat, I. and Remez, T. and Shaul, N. and Kreuk, F. and Chen, R. T. and Synnaeve, G. and Adi, Y. and Lipman, Y.},
  title = {Discrete flow matching},
  booktitle = {NeurIPS},
  year = {2024}
}

@article{ref39,
  author = {Chen, Z. and Wang, W. and Tian, H. and Ye, S. and Gao, Z. and Cui, E. and Tong, W. and Hu, K. and Luo, J. and Ma, Z. and others},
  title = {How far are we to gpt-4v? closing the gap to commercial multimodal models with open-source suites},
  journal = {arXiv preprint arXiv:2404.16821},
  year = {2024}
}

@article{ref40,
  author = {Chen, Z. and Wang, W. and Cao, Y. and Liu, Y. and Gao, Z. and Cui, E. and Zhu, J. and Ye, S. and Tian, H. and Liu, Z. and others},
  title = {Expanding performance boundaries of open-source multimodal models with model, data, and test-time scaling},
  journal = {arXiv preprint arXiv:2412.05271},
  year = {2024}
}

@article{ref41,
  author = {Lu, J. and Batra, D. and Parikh, D. and Lee, S.},
  title = {Vilbert: Pretraining taskagnostic visiolinguistic representations for vision-and-language tasks},
  journal = {Advances in neural information processing systems},
  volume = {32},
  year = {2019}
}

@article{ref42,
  author = {Li, L. H. and Yatskar, M. and Yin, D. and Hsieh, C.-J. and Chang, K.-W.},
  title = {Visualbert: A simple and performant baseline for vision and language},
  journal = {arXiv preprint arXiv:1908.03557},
  year = {2019}
}

@inproceedings{ref43,
  author = {Chen, Y.-C. and Li, L. and Yu, L. and El Kholy, A. and Ahmed, F. and Gan, Z. and Cheng, Y. and Liu, J.},
  title = {Uniter: Universal image-text representation learning},
  booktitle = {European conference on computer vision},
  publisher = {Springer},
  pages = {104--120},
  year = {2020}
}

@article{ref44,
  author = {Chen, J. and Huang, Y. and Lv, T. and Cui, L. and Chen, Q. and Wei, F.},
  title = {Textdiffuser: Diffusion models as text painters},
  journal = {Advances in Neural Information Processing Systems},
  volume = {36},
  year = {2024}
}

@misc{ref45,
  author = {{S. AI} and {LAION}},
  title = {Renderedtext},
  howpublished = {\url{https://huggingface.co/datasets/wendlerc/RenderedText}},
  year = {2023}
}

@inproceedings{ref46,
  author = {Nichol, A. Q. and Dhariwal, P. and Ramesh, A. and Shyam, P. and Mishkin, P. and Mcgrew, B. and Sutskever, I. and Chen, M.},
  title = {Glide: Towards photorealistic image generation and editing with text-guided diffusion models},
  booktitle = {International Conference on Machine Learning},
  publisher = {PMLR},
  pages = {16784--16804},
  year = {2022}
}

@article{ref47,
  author = {Saharia, C. and Chan, W. and Saxena, S. and Li, L. and Whang, J. and Denton, E. L. and Ghasemipour, K. and Gontijo Lopes, R. and Karagol Ayan, B. and Salimans, T. and others},
  title = {Photorealistic text-to-image diffusion models with deep language understanding},
  journal = {Advances in neural information processing systems},
  volume = {35},
  pages = {36479--36494},
  year = {2022}
}

@article{ref48,
  author = {Salimans, T. and Karpathy, A. and Chen, X. and Kingma, D. P.},
  title = {Pixelcnn++: Improving the pixelcnn with discretized logistic mixture likelihood and other modifications},
  journal = {arXiv preprint arXiv:1701.05517},
  year = {2017}
}

@inproceedings{ref49,
  author = {Reed, S. and Oord, A. and Kalchbrenner, N. and Colmenarejo, S. G. and Wang, Z. and Chen, Y. and Belov, D. and Freitas, N.},
  title = {Parallel multiscale autoregressive density estimation},
  booktitle = {International conference on machine learning},
  publisher = {PMLR},
  pages = {2912--2921},
  year = {2017}
}

@article{ref50,
  author = {Van Den Oord, A. and Vinyals, O. and others},
  title = {Neural discrete representation learning},
  journal = {Advances in neural information processing systems},
  volume = {30},
  year = {2017}
}

@article{ref51,
  author = {Razavi, A. and Van den Oord, A. and Vinyals, O.},
  title = {Generating diverse high-fidelity images with vq-vae-2},
  journal = {Advances in neural information processing systems},
  volume = {32},
  year = {2019}
}

@article{ref52,
  author = {Yu, J. and Li, X. and Koh, J. Y. and Zhang, H. and Pang, R. and Qin, J. and Ku, A. and Xu, Y. and Baldridge, J. and Wu, Y.},
  title = {Vector-quantized image modeling with improved vqgan},
  journal = {arXiv preprint arXiv:2110.04627},
  year = {2021}
}

@inproceedings{ref53,
  author = {Cao, S. and Yin, Y. and Huang, L. and Liu, Y. and Zhao, X. and Zhao, D. and Huang, K.},
  title = {Efficient-vqgan: Towards high-resolution image generation with efficient vision transformers},
  booktitle = {Proceedings of the IEEE/CVF International Conference on Computer Vision},
  pages = {7368--7377},
  year = {2023}
}

@article{ref54,
  author = {Yu, Q. and Weber, M. and Deng, X. and Shen, X. and Cremers, D. and Chen, L.-C.},
  title = {An image is worth 32 tokens for reconstruction and generation},
  journal = {Advances in Neural Information Processing Systems},
  volume = {37},
  pages = {128940--128966},
  year = {2024}
}

@article{ref55,
  author = {Zhu, L. and Wei, F. and Lu, Y. and Chen, D.},
  title = {Scaling the codebook size of vqgan to 100,000 with a utilization rate of 99\%},
  journal = {arXiv preprint arXiv:2406.11837},
  year = {2024}
}

@article{ref56,
  author = {Pang, Y. and Jin, P. and Yang, S. and Lin, B. and Zhu, B. and Tang, Z. and Chen, L. and Tay, F. E. and Lim, S.-N. and Yang, H. and others},
  title = {Next patch prediction for autoregressive visual generation},
  journal = {arXiv preprint arXiv:2412.15321},
  year = {2024}
}

@article{ref57,
  author = {Ren, S. and Ma, S. and Sun, X. and Wei, F.},
  title = {Next block prediction: Video generation via semi-auto-regressive modeling},
  journal = {arXiv preprint arXiv:2502.07737},
  year = {2025}
}

@article{ref58,
  author = {He, Y. and He, Y. and He, S. and Chen, F. and Zhou, H. and Zhang, K. and Zhuang, B.},
  title = {Neighboring autoregressive modeling for efficient visual generation},
  journal = {arXiv preprint arXiv:2503.10696},
  year = {2025}
}

@article{ref59,
  author = {Wang, Y. and Ren, S. and Lin, Z. and Han, Y. and Guo, H. and Yang, Z. and Zou, D. and Feng, J. and Liu, X.},
  title = {Parallelized autoregressive visual generation},
  journal = {arXiv preprint arXiv:2412.15119},
  year = {2024}
}

@article{ref60,
  author = {Tian, K. and Jiang, Y. and Yuan, Z. and Peng, B. and Wang, L.},
  title = {Visual autoregressive modeling: Scalable image generation via nextscale prediction},
  journal = {Advances in neural information processing systems},
  volume = {37},
  pages = {84839--84865},
  year = {2024}
}

@article{ref61,
  author = {Ren, S. and Yu, Q. and He, J. and Shen, X. and Yuille, A. and Chen, L.-C.},
  title = {Flowar: Scale-wise autoregressive image generation meets flow matching},
  journal = {arXiv preprint arXiv:2412.15205},
  year = {2024}
}

@article{ref63,
  author = {Schölkopf, B. and others},
  title = {Toward causal representation learning},
  journal = {Proceedings of the IEEE},
  volume = {109},
  number = {5},
  pages = {612--634},
  year = {2021}
}

@inproceedings{ref64,
  author = {Niu, Y. and Tang, K. and Zhang, H. and Lu, Z. and Hua, X. and Wen, J. R.},
  title = {Counterfactual VQA: A cause-and-effect look at language bias},
  booktitle = {Proceedings of the IEEE/CVF conference on computer vision and pattern recognition},
  pages = {12700--12710},
  year = {2021}
}

@inproceedings{ref65,
  author = {Agarwal, V. and Shetty, R. and Fritz, M.},
  title = {Towards Causal VQA: Revealing and Reducing Spurious Correlations by Invariant and Covariant Semantic Editing},
  booktitle = {CVPR},
  pages = {9690--9698},
  year = {2020}
}

@inproceedings{ref66,
  author = {Zhu, J.-Y. and Park, T. and Isola, P. and Efros, A. A.},
  title = {Unpaired image-to-image translation using cycle-consistent adversarial networks},
  booktitle = {Proceedings of the IEEE international conference on computer vision},
  pages = {2223--2232},
  year = {2017}
}

@inproceedings{ref67,
  author = {Qiao, T. and others},
  title = {MirrorGAN: Learning Text-to-Image Generation by Redescription},
  booktitle = {CVPR},
  pages = {1505--1514},
  year = {2019}
}

@article{ref68,
  author = {{Chameleon Team}},
  title = {Chameleon: Mixed-modal early-fusion foundation models},
  journal = {arXiv preprint arXiv:2405.09818},
  year = {2024}
}

@article{ref69,
  author = {Xie, J. and Mao, W. and Bai, Z. and others},
  title = {Show-o: One Single Transformer to Unify Multimodal Understanding and Generation},
  journal = {arXiv preprint arXiv:2408.12528},
  year = {2024}
}

@inproceedings{ref70,
  author = {Liu, H. and Li, C. and Wu, Q. and Lee, Y. J.},
  title = {Visual instruction tuning},
  booktitle = {Proceedings of the Conference on Neural Information Processing Systems},
  year = {2023}
}

@article{ref71,
  author = {Bai, J. and Bai, S. and Yang, S. and Wang, S. and Tan, S. and Wang, P. and Lin, J. and Zhou, C. and Zhou, J.},
  title = {Qwen-VL: A versatile vision-language model for understanding, localization, text reading, and beyond},
  journal = {arXiv preprint arXiv:2308.12966},
  year = {2023}
}

@article{ref72,
  author = {Li, B. and Zhang, Y. and Guo, D. and Zhang, R. and Li, F. and Zhang, H. and Zhang, K. and Zhang, P. and others},
  title = {LLaVA-OneVision: Easy visual task transfer},
  journal = {arXiv preprint arXiv:2408.03326},
  year = {2024}
}

@article{ref73,
  author = {Wu, S. and Fei, H. and Qu, L. and Ji, W. and Chua, T. S.},
  title = {NExT-GPT: Any-to-any multimodal LLM},
  journal = {arXiv preprint arXiv:2309.05519},
  year = {2023}
}

@article{ref74,
  author = {Ge, Y. and Zhao, S. and Zhu, J. and Ge, Y. and Yi, K. and Song, L. and Li, C. and Ding, D. and Shan, Y.},
  title = {SEED-X: Multimodal models with unified multi-granularity comprehension and generation},
  journal = {arXiv preprint arXiv:2404.14396},
  year = {2024}
}

@article{ref75,
  author = {Tong, S. and Fan, D. and Zhu, J. and Xiong, Y. and Chen, X. and Sinha, K. and Rabbat, M. and LeCun, Y. and others},
  title = {MetaMorph: Multimodal understanding and generation via instruction tuning},
  journal = {arXiv preprint arXiv:2412.14164},
  year = {2024}
}

@inproceedings{ref76,
  author = {Qu, L. and Zhang, J. and Wu, H. and Xing, J. and Xia, M. and Zhang, Y. and Shou, MZ},
  title = {TokenFlow: Unified image tokenizer for multimodal understanding and generation},
  booktitle = {Proceedings of the IEEE/CVF Conference on Computer Vision and Pattern Recognition},
  year = {2025}
}

@inproceedings{ref77,
  author = {Wang, C. and Ma, H. and Wang, J. and Zhu, J. and Guan, Y. and Zeng, Y. and Gao, L. and Shou, MZ},
  title = {ILLUME: Illuminating your LLMs to see, draw, and self-enhance},
  booktitle = {Proceedings of the IEEE/CVF International Conference on Computer Vision},
  year = {2025}
}

@article{ref78,
  author = {Lin, B. and Li, Z. and Cheng, X. and Niu, Y. and Ye, Y. and He, X. and Yuan, S. and Song, L.},
  title = {Emerging properties in unified multimodal pretraining},
  journal = {arXiv preprint arXiv:2505.14683},
  year = {2025}
}

@article{ref79,
  author = {Xie, J. and Song, W. and Xing, J. and Wu, H. and Wang, J. and Qu, L. and Xing, J. and Shou, MZ},
  title = {Show-o: One single transformer to unify multimodal understanding and generation},
  journal = {arXiv preprint arXiv:2408.12528},
  year = {2024}
}

@article{ref80,
  author = {Ma, Y. and Liu, X. and Chen, X. and Liu, W. and Wu, C. and Wu, Z. and Pan, Z. and others},
  title = {JanusFlow: Harmonizing autoregression and rectified flow for unified multimodal understanding and generation},
  journal = {arXiv preprint arXiv:2411.07975},
  year = {2024}
}

@inproceedings{ref81,
  author = {Li, H. and Tian, C. and Shao, J. and Zhu, X. and Wang, Z. and Zhu, J. and Dou, W. and others},
  title = {SynerGen-VL: Towards synergistic image understanding and generation with vision experts and token folding},
  booktitle = {Proceedings of the IEEE/CVF Conference on Computer Vision and Pattern Recognition},
  year = {2025}
}

@article{ref82,
  author = {{DeepSeek-AI} and Chen, C. and Gao, Q. and Liu, Z. and Liu, W. and Chen, K.},
  title = {Janus-Pro: Unified multimodal understanding and generation with optimized training strategy},
  journal = {arXiv preprint arXiv:2501.17811},
  year = {2025}
}

@article{ref83,
  author = {Xie, J. and Yang, Z. and Shou, MZ},
  title = {Show-o2: Improved native unified multimodal models},
  journal = {arXiv preprint arXiv:2506.15564},
  year = {2025}
}

@article{ref84,
  author = {Wang, X. and Wang, Z. and Zhang, J. and Peng, S. and Huang, Z. and Peng, S. and Zhou, Y. and Huang, B.},
  title = {Emu3: Next-token prediction is all you need},
  journal = {arXiv preprint arXiv:2409.18869},
  year = {2024}
}

@article{ref85,
  author = {Wu, Y. and Zhang, Z. and Chen, J. and Tang, J. and others},
  title = {VILA-U: a unified foundation model integrating visual understanding and generation},
  journal = {arXiv preprint arXiv:2409.04429},
  year = {2024}
}

@inproceedings{ref86,
  author = {Xie, J. and Wang, J. and Xing, J. and Shou, MZ},
  title = {MUSE-VL: Modeling unified VLM through semantic discrete encoding},
  booktitle = {Proceedings of the IEEE/CVF International Conference on Computer Vision},
  year = {2025}
}

@article{ref87,
  author = {Zhou, C. and Yu, J. C. and Wang, N. M. and Gelada, C. and Lenc, K. and Gu, M. and Dai, B.},
  title = {Liquid: Language models are scalable and unified multi-modal generators},
  journal = {International Journal of Computer Vision},
  year = {2025}
}

@article{ref88,
  author = {Zhang, L. and Liu, K. and Xie, T. and Huang, J. and Yu, Z. and Busch, K.},
  title = {MoGA-ETA: Generalized face anti-spoofing with enhanced text guidance and alignment},
  journal = {IEEE Transactions on Information Forensics and Security},
  year = {2026}
}

@article{ref90,
  author = {Pan, X. and Shukla, S. N. and Singh, A. and Zhao, Z. and Mishra, S. K. and Juefei-Xu, F.},
  title = {Transfer between modalities with MetaQueries},
  journal = {arXiv preprint arXiv:2504.06256},
  year = {2025}
}

@inproceedings{ref91,
  author = {Zhou, C. and Yu, J. C. and Wang, N. M. and Gelada, C. and Lenc, K. and Gu, M. and Dai, B.},
  title = {Transfusion: Predict the next token and diffuse images with one multi-modal model},
  booktitle = {Proceedings of the International Conference on Learning Representations},
  year = {2025}
}

@inproceedings{ref92,
  author = {Li, Z. and Li, H. and Shi, Y. and Farimani, A. B. and Moura, J. M.},
  title = {Dual diffusion for unified image generation and understanding},
  booktitle = {Proceedings of the IEEE/CVF Conference on Computer Vision and Pattern Recognition},
  year = {2025}
}

@article{ref93,
  author = {{OpenAI} and Achiam, J. and Adler, S. and Agarwal, S. and Ahmad, L. and Akkaya, I. and Aleman, F. L. and Almeida, D. and Altenschmidt, J. and Altman, S. and others},
  title = {GPT-4 technical report},
  journal = {arXiv preprint arXiv:2303.08774},
  year = {2023}
}

@article{ref94,
  author = {{OpenAI}},
  title = {GPT-4o system card},
  journal = {arXiv preprint arXiv:2410.21276},
  year = {2024}
}

@article{ref95,
  author = {{Gemini Team, Google}},
  title = {Gemini 1.5: Unlocking multimodal understanding across millions of tokens of context},
  journal = {arXiv preprint arXiv:2403.05530},
  year = {2024}
}

@inproceedings{ref96,
  author = {Lin, J. and Yin, H. and Ping, W. and Lu, Y. and Molchanov, S. and Tao, A. and Hu, H. and Han, S.},
  title = {VILA: On pre-training for visual language models},
  booktitle = {Proceedings of the IEEE/CVF Conference on Computer Vision and Pattern Recognition},
  year = {2024}
}

@article{ref97,
  author = {Xu, L. and Zhao, Z. and Song, W. and Xing, J. and Wu, H. and Wang, J. and Qu, L. and Xing, J. and Shou, MZ},
  title = {PLLaVA: Parameter-free LLaVA extension from images to videos for video dense captioning},
  journal = {arXiv preprint arXiv:2404.16994},
  year = {2024}
}

@article{ref98,
  author = {Zhang, P. and Dong, X. and Cao, Y. and Jin, S. and Zang, Y. and Chen, K. and Wang, J.},
  title = {Long context transfer from language to vision},
  journal = {arXiv preprint arXiv:2406.16852},
  year = {2024}
}

@article{ref99,
  author = {Zhang, P. and Dong, X. and Zang, Y. and Cao, Y. and Chen, C. and Pan, L. and Chen, K. and Wang, J.},
  title = {InternLM-XComposer2.5: A comprehensive multimodal system for long-term streaming video and audio interactions},
  journal = {arXiv preprint arXiv:2412.09596},
  year = {2024}
}

@article{ref100,
  author = {Cheng, Z. and Pu, S. and Wang, J. and others},
  title = {VideoLLaMA 2: Advancing video understanding with audio-visual integration},
  journal = {arXiv preprint arXiv:2406.11896},
  year = {2024}
}

@article{ref101,
  author = {Li, Z. and Zhang, J. and Lin, Q. and Xiong, J. and Long, Y. and Deng, X. and Zhang, Y. and Liu, X. and Huang, M. and Xiao, Z. and others},
  title = {Hunyuan-DiT: A powerful multi-resolution diffusion transformer with fine-grained Chinese understanding},
  journal = {arXiv preprint arXiv:2405.08748},
  year = {2024}
}

@article{ref102,
  author = {Li, D. and Kamko, A. and Akhgari, E. and Sabet, A. and Xu, L. and others},
  title = {Playground v2.5: Three insights towards enhancing aesthetic quality in text-to-image generation},
  journal = {arXiv preprint arXiv:2402.17245},
  year = {2024}
}

@inproceedings{ref103,
  author = {Chen, J. and Ge, C. and Xie, E. and Wu, Y. and Yao, L. and Ren, X. and Wang, Z. and Luo, P. and Lu, H. and Li, Z.},
  title = {PixArt-Sigma: Weak-to-strong training of diffusion transformer for 4K text-to-image generation},
  booktitle = {Proceedings of the European Conference on Computer Vision},
  year = {2024}
}

@misc{ref104,
  author = {Betker, J. and Goh, G. and Jing, L. and Brooks, T. and Wang, J. and Li, L. and Ouyang, L. and Zhuang, J. and Lee, J. and Guo, Y. and others},
  title = {Improving image generation with better captions},
  howpublished = {OpenAI Technical Report},
  year = {2023}
}

@article{ref105,
  author = {Wang, X. and Wang, Z. and Zhang, J. and Peng, S. and Huang, Z. and Peng, S. and Zhou, Y. and Huang, B.},
  title = {Multimodal learning with next-token prediction for large multimodal models},
  journal = {Nature},
  year = {2026}
}

@misc{ref106,
  author = {{Stability AI}},
  title = {Stable Diffusion 3.5 technical report},
  howpublished = {Stability AI},
  year = {2024}
}

@misc{ref107,
  author = {{Black Forest Labs}},
  title = {FLUX.1: Flow matching for high-resolution image synthesis},
  howpublished = {Black Forest Labs Technical Report},
  year = {2024}
}

@article{ref108,
  author = {Xie, E. and Chen, J. and Yu, H. and Ge, Z. and Yao, L. and Li, Z. and Luo, P.},
  title = {SANA: Efficient high-resolution image synthesis with linear diffusion transformer},
  journal = {arXiv preprint arXiv:2410.10629},
  year = {2024}
}

@inproceedings{ref109,
  author = {Qin, Q. and Zhuo, L. and Xin, Y. and Du, R. and Li, Z. and Fu, B. and Lu, Y. and Li, X. and Liu, D. and Zhu, X. and Beddow, W. and others},
  title = {Lumina-Image 2.0: A unified and efficient image generative framework},
  booktitle = {Proceedings of the IEEE/CVF International Conference on Computer Vision},
  year = {2025}
}

@article{ref110,
  author = {Cai, Q. and Yang, H. and Pu, Y. and Qiu, B. and Paul, S. and Liao, Y. and Liu, Y. and others},
  title = {HiDream-I1: A high-efficient image generative foundation model with sparse diffusion transformer},
  journal = {arXiv preprint arXiv:2505.22705},
  year = {2025}
}

@article{ref111,
  author = {Xue, L. and Shu, M. and Awadalla, A. and Wang, J. and Gui, A. and Yan, S. and Sun, C. and others},
  title = {xGen-MM (BLIP-3): A family of open large multimodal models},
  journal = {arXiv preprint arXiv:2408.08872},
  year = {2024}
}

@article{ref112,
  author = {Wu, C. and Zheng, P. and Yan, R. and Xiao, S. and Luo, X. and Wang, Y. and Li, W. and Jiang, X. and Liu, Y. and others},
  title = {OmniGen2: Exploration to advanced multimodal generation},
  journal = {arXiv preprint arXiv:2506.18871},
  year = {2025}
}

@article{ref113,
  author = {Wang, J. and Yuan, H. and Chen, D. and Zhang, Y. and Wang, X. and Zhang, S.},
  title = {ModelScope text-to-video technical report},
  journal = {arXiv preprint arXiv:2308.06571},
  year = {2023}
}

@article{ref114,
  author = {Wang, Y. and Chen, X. and Ma, X. and Zhou, S. and Huang, Z. and Yi, Y. and others},
  title = {LaVie: High-quality video generation with cascaded latent diffusion models},
  journal = {arXiv preprint arXiv:2309.15103},
  year = {2023}
}

@article{ref115,
  author = {Lin, B. and Ge, Y. and Cheng, X. and others},
  title = {Open-Sora Plan: Open-source large video generation model},
  journal = {arXiv preprint arXiv:2412.00131},
  year = {2024}
}

@article{ref116,
  author = {Zhang, D. and Huang, J. and Pan, S. and others},
  title = {Show-1: Marrying pixel and latent diffusion models for text-to-video generation},
  journal = {arXiv preprint arXiv:2309.15818},
  year = {2023}
}

@inproceedings{ref117,
  author = {Guo, Y. and Yang, C. and Rao, A. and Wang, Y. and Qiao, Y. and Lin, D. and Dai, B.},
  title = {AnimateDiff: Animate your personalized text-to-image diffusion models without specific tuning},
  booktitle = {Proceedings of the International Conference on Learning Representations},
  year = {2024}
}

@inproceedings{ref118,
  author = {Chen, H. and Zhang, Y. and Cun, X. and Xia, M. and Wang, X. and Weng, C. and Shan, Y.},
  title = {VideoCrafter2: Overcoming data limitations for high-quality video diffusion models},
  booktitle = {Proceedings of the IEEE/CVF Conference on Computer Vision and Pattern Recognition},
  year = {2024}
}

@article{ref119,
  author = {Yang, Z. and Teng, J. and Zheng, W. and Ding, M. and Huang, S. and others},
  title = {CogVideoX: Text-to-video diffusion models with an expert transformer},
  journal = {arXiv preprint arXiv:2408.06072},
  year = {2024}
}

@article{ref120,
  author = {Ma, G. and Duan, N. and Chen, X. and Wan, C. and Huang, H. and others},
  title = {Step-Video-T2V technical report: The practice, challenges, and future of video foundation model},
  journal = {arXiv preprint arXiv:2502.10248},
  year = {2025}
}

@misc{ref121,
  author = {{Runway}},
  title = {Introducing Gen-3 Alpha},
  howpublished = {Runway Research Blog},
  year = {2024}
}

@inproceedings{ref122,
  author = {Xing, J. and Xia, M. and Zhang, Y. and Chen, H. and Yu, W. and Liu, H. and others},
  title = {DynamiCrafter: Animating open-domain images with video diffusion priors},
  booktitle = {Proceedings of the European Conference on Computer Vision},
  year = {2024}
}

@inproceedings{ref123,
  author = {Chen, X. and Wang, Y. and Zhang, L. and Zhuang, S. and Ma, X. and Yu, J. and Wang, Y. and Lin, D. and Qiao, Y. and Liu, Z.},
  title = {SEINE: Short-to-long video diffusion model for generative transition and prediction},
  booktitle = {Proceedings of the International Conference on Learning Representations},
  year = {2024}
}

@article{ref124,
  author = {Zhang, S. and Wang, J. and Zhang, Y. and Zhao, K. and Yuan, H. and Qin, Z. and Wang, X. and Zhao, D. and Zhou, J.},
  title = {I2VGen-XL: High-quality image-to-video synthesis via cascaded diffusion models},
  journal = {arXiv preprint arXiv:2311.05127},
  year = {2023}
}

@inproceedings{ref125,
  author = {Lei, G. and Huang, J. and Pan, S. and others},
  title = {Animate Anything: Consistent and controllable animation for video generation},
  booktitle = {Proceedings of the IEEE/CVF Conference on Computer Vision and Pattern Recognition},
  year = {2025}
}

@article{ref126,
  author = {Ren, W. and Yang, H. and Zhang, G. and Wei, C. and Du, X. and Huang, W. and Chen, W.},
  title = {ConsistI2V: Enhancing visual consistency for image-to-video generation},
  journal = {Transactions on Machine Learning Research},
  year = {2024}
}

@article{ref127,
  author = {Wu, H. and Zhang, E. and Liao, L. and Chen, C. and others},
  title = {VideoCrafter-I2V: A pioneering model in image-to-video generation},
  journal = {arXiv preprint arXiv:2310.15731},
  year = {2023}
}

@article{ref128,
  author = {Blattmann, A. and Dockhorn, T. and Kulal, S. and Podell, D. and Entezari, R. and others},
  title = {Stable Video Diffusion: Scaling latent video diffusion models to large datasets},
  journal = {arXiv preprint arXiv:2311.15127},
  year = {2023}
}

@article{ref129,
  author = {Liu, H. and Liu, S. and Zhou, Z. and Xu, M. and Xie, Y. and Han, X. and P{\'e}rez, J. C. and Liu, D. and Kahatapitiya, K. and others},
  title = {MarDini: Masked autoregressive diffusion for video generation at scale},
  journal = {Transactions on Machine Learning Research},
  year = {2024}
}

@article{ref130,
  author = {Zhang, L. and Liu, K. and Xie, T. and Huang, J. and Yu, Z. and Busch, K.},
  title = {MoGA-ETA: Generalized face anti-spoofing with enhanced text guidance and alignment},
  journal = {IEEE Transactions on Information Forensics and Security},
  year = {2026}
}

@inproceedings{ref131,
  author = {Esser, P. and Kulal, S. and Blattmann, A. and Entezari, R. and M{\"u}ller, J. and Saini, H. and Levi, Y. and Lorenz, D. and Sauer, A. and Boesel, F. and Podell, D. and Dockhorn, T. and English, Z. and Lacey, K. and Goodwin, A. and Marek, Y. and Rombach, R.},
  title = {Scaling rectified flow transformers for high-resolution image synthesis},
  booktitle = {Proceedings of the IEEE/CVF Conference on Computer Vision and Pattern Recognition},
  year = {2024}
}

\appendix

\section{Additional Implementation Details}

\label{app:implementation}

This section provides detailed implementation settings of C3-UniMM. Our model is built upon Show-o2, and we retain its native unified multimodal modeling design while adding the proposed Structured Latent Causal Graph (SLCG), Unified Decoding Space (UDS), Super Alignment, causal intervention, and cycle-consistency learning modules.

\subsection{Backbone Architecture}

\paragraph{Base model.}

C3-UniMM follows the native unified architecture of Show-o2. The backbone contains a text tokenizer and embedding layer, a 3D causal VAE encoder/decoder for image and video inputs, semantic layers, a low-level projector, a spatial-temporal fusion module, a pretrained language model, a language modeling head, and a flow matching head.

Given an interleaved text-image-video input, text tokens are embedded by the text embedding layer, while images and videos are encoded into continuous visual latents by the 3D causal VAE encoder. The visual latents are processed by two paths: semantic layers extract high-level semantic features, while the projector preserves low-level visual details. The two paths are fused by the spatial-temporal fusion module to form unified visual representations. The resulting visual representations and text embeddings are arranged into a single interleaved sequence and fed into the language model.

Following Show-o2, the language modeling head is used for autoregressive text token prediction, and the flow matching head is used for image and video latent generation. The generated image/video latents are finally decoded by the frozen 3D causal VAE decoder.

\paragraph{Model scales.}

We instantiate two model sizes:
\[
\text{C3-UniMM-1.5B}, \qquad \text{C3-UniMM-7B}.
\]
The 1.5B model uses Qwen2.5-1.5B-Instruct as the base language model, and the 7B model uses Qwen2.5-7B-Instruct. Unless otherwise specified, the 3D causal VAE is initialized from the Show-o2 backbone and kept frozen throughout training.

\subsection{Structural Slot Adapter}

C3-UniMM introduces a structural slot adapter on top of the unified hidden representations produced by the Show-o2 backbone. Let $H_m\in\mathbb{R}^{L_m\times d_h}$ denote the hidden sequence of modality $m$, where $L_m$ is the sequence length and $d_h$ is the hidden dimension of the base language model. We first project $H_m$ into a shared structural space:
\[
\tilde{H}_m = \mathrm{MLP}_{\mathrm{in}}(H_m),
\qquad
\tilde{H}_m\in\mathbb{R}^{L_m\times d_s},
\]
where $d_s=1024$ is the structural hidden dimension.

We then apply Slot Attention to extract $K=16$ object-centric slots:
\[
S_m=\mathrm{SlotAttn}(\tilde{H}_m)
=
\{s_{m,1},\dots,s_{m,K}\},
\qquad
s_{m,i}\in\mathbb{R}^{d_s}.
\]
The Slot Attention module uses 3 iterative refinement steps and 8 attention heads. For image inputs, slots are encouraged to represent object-level regions. For video inputs, they capture spatio-temporal entities or events. For text inputs, slots are extracted from content-token hidden states and represent semantic entities, attributes, or relations.

\subsection{Structured Latent Causal Graph Construction}

Given the slot set $S_m$, we construct a directed latent graph
\[
G_m=(S_m,A_m),
\]
where $A_m\in[0,1]^{K\times K}$ is a directed adjacency matrix. We compute $A_m$ as
\[
A_m=
\sigma
\left(
\frac{(S_mW_q)(S_mW_k)^\top}{\sqrt{d_s}}
\right)
\odot(1-I),
\]
where $W_q,W_k\in\mathbb{R}^{d_s\times d_s}$ are learnable projections, $I$ is the identity matrix, and $(1-I)$ removes self-loops. The entry $A_{m,ji}$ represents the dependency strength from slot $j$ to slot $i$.

To prevent the graph from degenerating into an unconstrained correlation matrix, we add graph regularization:
\[
\mathcal{L}_{\mathrm{graph}}
=
\lambda_{\mathrm{dag}}h(A_m)
+
\lambda_{\mathrm{sp}}\|A_m\|_1
+
\lambda_{\mathrm{rel}}\mathcal{L}_{\mathrm{rel}},
\]
where
\[
h(A_m)
=
\mathrm{tr}
\left(
\exp(A_m\odot A_m)
\right)-K.
\]
The acyclicity penalty $h(A_m)$ encourages approximately directed acyclic structures, $\|A_m\|_1$ encourages sparse dependencies, and $\mathcal{L}_{\mathrm{rel}}$ is used when relation annotations are available. In our implementation, relation supervision is applied to structured datasets such as CLEVR and Visual Genome, while it is omitted for ordinary image-text and video-text pairs.

\subsection{Unified Decoding Space}

The graph $G_m$ is mapped into a Unified Decoding Space before conditioning the Show-o2 language and flow heads. We use a lightweight graph transformer to aggregate node and edge information:
\[
H_m^G=\mathrm{GraphTrans}(S_m,A_m).
\]
The graph-level representation is obtained by mean pooling:
\[
\bar{G}_m=\mathrm{MeanPool}(H_m^G).
\]
We then project it into the UDS:
\[
z_m=g(G_m)=\mathrm{MLP}_{g}(\bar{G}_m),
\qquad
z_m\in\mathbb{R}^{1024}.
\]
To inject structural information into the Show-o2 backbone, $z_m$ is further projected to the hidden size of the base language model and expanded into $N_p$ structural prefix tokens:
\[
P_m=\mathrm{MLP}_{p}(z_m)\in\mathbb{R}^{N_p\times d_h}.
\]
We set $N_p=8$ by default. These structural prefix tokens are concatenated with the original interleaved text-visual sequence. During understanding tasks, they provide object-relation information to the language head. During image and video generation, they condition the flow head so that visual latent generation is guided by structured semantic information.

\subsection{Super Alignment}

For paired multimodal samples, such as image-caption, video-caption, or image-video-text triples, we extract latent graphs from different modalities and align them at representation, structure, and mechanism levels.

Because slot ordering may differ across modalities, we first compute a soft permutation matrix $P_{mn}$ using Sinkhorn matching over slot similarities:
\[
\tilde{S}_n=P_{mn}S_n,
\qquad
\tilde{A}_n=P_{mn}A_nP_{mn}^{\top}.
\]
The Super Alignment loss is
\[
\mathcal{L}_{\mathrm{SA}}
=
\mathcal{L}_{\mathrm{rep}}
+
\alpha_{\mathrm{str}}\mathcal{L}_{\mathrm{str}}
+
\alpha_{\mathrm{mech}}\mathcal{L}_{\mathrm{mech}},
\]
where
\[
\mathcal{L}_{\mathrm{rep}}
=
\|\bar{G}_m-\bar{G}_n\|_2^2,
\]
and
\[
\mathcal{L}_{\mathrm{str}}
=
\|A_m-\tilde{A}_n\|_F^2.
\]
For mechanism alignment, we parameterize each node mechanism with a shared mechanism encoder $M_\eta$. The parent context of node $i$ is computed as
\[
c_{m,i}
=
\sum_{j\neq i}
A_{m,ji}\psi(s_{m,j}),
\]
where $\psi(\cdot)$ is a two-layer MLP. The mechanism representation is
\[
r_{m,i}
=
M_\eta([s_{m,i},c_{m,i}]).
\]
The mechanism-level loss is
\[
\mathcal{L}_{\mathrm{mech}}
=
\sum_{i=1}^{K}
\|r_{m,i}-\tilde{r}_{n,i}\|_2^2.
\]
This makes mechanism alignment directly trainable instead of comparing unspecified abstract functions.

\subsection{Cycle Consistency and Intervention Training}

\paragraph{Cycle consistency.}

Given an input $x_m$ from modality $m$, we first infer its graph $G_m$, generate a target modality sample $\hat{x}_n$, and then re-encode $\hat{x}_n$ into a graph $\hat{G}_n$. The structural cycle loss is
\[
\mathcal{L}_{\mathrm{cyc}}^{G}
=
d_{\mathcal{G}}(\hat{G}_n,G_m),
\]
where
\[
d_{\mathcal{G}}(\hat{G}_n,G_m)
=
\| \hat{S}_n-S_m \|_2^2
+
\| \hat{A}_n-A_m \|_F^2.
\]
When reconstruction to the original modality is available, we also add an observation-level cycle loss:
\[
\mathcal{L}_{\mathrm{cyc}}^{x}
=
d_x(\hat{x}_m,x_m).
\]
For text, $d_x$ is implemented as token-level cross-entropy. For images and videos, it is computed in the VAE latent space using an $\ell_2$ distance.

The final cycle loss is
\[
\mathcal{L}_{\mathrm{cyc}}
=
\mathcal{L}_{\mathrm{cyc}}^{G}
+
\lambda_x
\mathcal{L}_{\mathrm{cyc}}^{x}.
\]

\paragraph{Intervention training.}

To improve controllability, we randomly select a slot or an edge and perform latent intervention. For node intervention, we replace slot $s_k$ with an edited value $v$:
\[
G_m'=do(G_m,s_k=v).
\]
The edited graph is decoded into target modality $n$:
\[
\tilde{x}_n=D_n(g(G_m')).
\]
We then re-encode $\tilde{x}_n$ into a graph $\tilde{G}_n$ and minimize
\[
\mathcal{L}_{\mathrm{int}}
=
d_{\mathcal{G}}(\tilde{G}_n,G_m')
+
\lambda_{\mathrm{edit}}\mathcal{L}_{\mathrm{edit}}.
\]
For CLEVR, $\mathcal{L}_{\mathrm{edit}}$ is computed from ground-truth object attributes and relations. For Visual Genome and open-domain data, we use pretrained detectors, captioners, or VQA models to check whether the intended attribute or relation change is reflected in the generated output.

\subsection{Training Objective}

The base Show-o2\cite{ref83} objective consists of next-token prediction and flow matching:
\[
\mathcal{L}_{\mathrm{base}}
=
\alpha_{\mathrm{NTP}}\mathcal{L}_{\mathrm{NTP}}
+
\mathcal{L}_{\mathrm{FM}}.
\]
C3-UniMM augments this objective with structural graph learning, Super Alignment, cycle consistency, and intervention training:
\[
\mathcal{L}
=
\mathcal{L}_{\mathrm{base}}
+
\lambda_{\mathrm{graph}}\mathcal{L}_{\mathrm{graph}}
+
\lambda_{\mathrm{SA}}\mathcal{L}_{\mathrm{SA}}
+
\lambda_{\mathrm{cyc}}\mathcal{L}_{\mathrm{cyc}}
+
\lambda_{\mathrm{int}}\mathcal{L}_{\mathrm{int}}.
\]
Unless otherwise specified, we use
\[
\lambda_{\mathrm{graph}}=0.1,\quad
\lambda_{\mathrm{SA}}=0.5,\quad
\lambda_{\mathrm{cyc}}=0.2,\quad
\lambda_{\mathrm{int}}=0.2.
\]
For graph regularization, we set
\[
\lambda_{\mathrm{dag}}=1.0,\quad
\lambda_{\mathrm{sp}}=0.01.
\]
For Super Alignment, we set
\[
\alpha_{\mathrm{str}}=1.0,\quad
\alpha_{\mathrm{mech}}=0.5.
\]

\subsection{Training Schedule}

We follow the two-stage training recipe of Show-o2 and add structure-specific training objectives for C3-UniMM. The 3D causal VAE is frozen in all stages.

\paragraph{Stage 0: initialization.}

We initialize the text tokenizer, text embedding layer, 3D causal VAE, semantic layers, projector, spatial-temporal fusion module, language model, language head, and flow head from Show-o2. The newly introduced slot adapter, graph transformer, mechanism encoder, UDS projector, and structural prefix projector are randomly initialized.

\paragraph{Stage 1: visual generation warm-up.}

We train the projector, spatial-temporal fusion module, flow head, and newly added structural modules on image-text pairs. The language model is frozen at the beginning of this stage to preserve language knowledge. We use image resolution $432\times432$, context length 1024 for single image-text pairs, and train for 150K iterations. The base loss weight is set to $\alpha_{\mathrm{NTP}}=0.2$. Captions are dropped with probability 0.1 for classifier-free guidance.

\paragraph{Stage 1.5: high-quality generation refinement.}

We replace the generation data with high-quality image-text pairs and continue training for 40K iterations. In this stage, $\mathcal{L}_{\mathrm{graph}}$ and $\mathcal{L}_{\mathrm{SA}}$ are enabled for all paired samples. Relation supervision is enabled only for samples with relation annotations.

\paragraph{Stage 2: multimodal instruction tuning.}

We fine-tune the full model except the frozen VAE using multimodal instruction data, high-quality generation data, structured reasoning data, and video-text data. This stage lasts 35K iterations. We set $\alpha_{\mathrm{NTP}}=1.0$ to strengthen language and instruction-following ability. Cycle consistency and intervention losses are activated in this stage.

For video training, we sample 17 frames from each clip. We use either 480p or $432\times432$ resolution, and set the maximum context length to 7006 for video and mixed-modality inputs.

\paragraph{Stage 3: structural consistency tuning.}

Finally, we perform an additional structure-focused tuning stage. We oversample CLEVR, Visual Genome, GenEval-style compositional prompts, relation-heavy image-text samples, and video-text samples with explicit object or motion relations. This stage focuses on improving structural preservation, compositional generation, and intervention consistency. The base Show-o2 losses remain active, while $\mathcal{L}_{\mathrm{SA}}$, $\mathcal{L}_{\mathrm{cyc}}$, and $\mathcal{L}_{\mathrm{int}}$ are increased by a factor of 1.5. The total training length is approximately 300K iterations.

\subsection{Scaling from 1.5B to 7B}

For C3-UniMM-7B, we initialize the base language model from Qwen2.5-7B-Instruct and reuse the flow head and structural modules trained from the 1.5B model whenever dimensions permit. Since the 7B language model has a different hidden size, we add lightweight MLP adapters to map structural prefix tokens and flow hidden states to the 7B hidden dimension.

We first train only the newly initialized MLP adapters, structural prefix projector, slot adapter, projector, and spatial-temporal fusion module for 3K iterations with 2K warm-up steps. We then follow the same Stage-1 to Stage-3 training recipe as the 1.5B model.

\subsection{Optimization Hyperparameters}

All models are trained with AdamW and mixed precision. We use gradient clipping with a maximum norm of 1.0 and cosine learning-rate decay after warm-up. Unless otherwise specified, weight decay is set to 0.01.

\begin{table}[H]
\centering
\caption{Default hyperparameters of C3-UniMM.}
\label{tab:hyperparams}
\begin{tabular}{ll}
\toprule
\textbf{Hyperparameter} & \textbf{Value} \\
\midrule
Number of slots $K$ & 16 \\
Structural hidden dimension $d_s$ & 1024 \\
Slot Attention iterations & 3 \\
Slot Attention heads & 8 \\
Graph transformer layers & 2 \\
Graph transformer heads & 8 \\
Structural prefix tokens $N_p$ & 8 \\
Image resolution & $432\times432$ \\
Video frames & 17 \\
Image-text context length & 1024 \\
Video/mixed-modality context length & 7006 \\
Optimizer & AdamW \\
Weight decay & 0.01 \\
Gradient clipping & 1.0 \\
Training precision & Mixed precision \\
Total training iterations & 300K \\
\bottomrule
\end{tabular}
\end{table}

\subsection{Inference}

At inference time, text outputs are decoded autoregressively from the language head. Image and video outputs are generated by sampling visual latents through the flow head and decoding them with the frozen 3D causal VAE decoder.

For controllable generation, users can edit either node slots or graph edges before UDS projection. Node editing modifies object-level attributes or entities, while edge editing modifies relations between objects. The edited graph is then projected into the UDS and used to condition the flow head for structure-aware image or video generation.

\section{Supplementary Experimental Data}

\begin{table}[H]
\centering
\caption{Benchmark results comparing our C3-UniMM model against existing Gen-Only and Native Unified methods.}
\label{tab:benchmark-results-3}
\resizebox{\textwidth}{!}{
\begin{tabular}{llcccccccc}
\toprule
\textbf{Type} & \textbf{Method} & \textbf{\#Params} & \textbf{\#Data} & \textbf{Global} & \textbf{Entity} & \textbf{Attribute} & \textbf{Relation} & \textbf{Other} & \textbf{Overall$\uparrow$} \\ 
\midrule
\multirow{4}{*}{Gen Only} & Hunyuan-DiT \cite{ref101} & 1.5B & - & 84.59 & 80.59 & 88.01 & 74.36 & 86.41 & 78.87 \\
 & Playground v2.5 \cite{ref102} & - & - & 83.06 & 82.59 & 81.2 & 84.08 & 83.5 & 75.47 \\
 & PixArt-$\Sigma$ \cite{ref103} & - & - & 86.89 & 82.89 & 88.94 & 86.59 & 87.68 & 80.54 \\
 & DALL-E 3 \cite{ref104} & - & - & 90.97 & 89.61 & 88.39 & 90.58 & 89.83 & 83.5 \\ 
\midrule
\multirow{8}{*}{Native Unified} & SD3-Medium \cite{ref101} & 2B & - & 87.9 & 91.01 & 88.83 & 80.7 & 88.68 & 84.08 \\
 & Emu3-DPO \cite{ref105} & 8B & - & - & - & - & - & - & 81.6 \\
 & Janus-Pro \cite{ref82} & 7B & 144M & 86.9 & 88.9 & 89.4 & 89.32 & 89.48 & 84.19 \\
 & Moga \cite{ref88} & 7B & - & 82.37 & 90.03 & 88.26 & 93.18 & 85.4 & 84.33 \\
 & Show-o2 \cite{ref83} & 1.5B & 66M & 87.53 & 90.38 & 91.34 & 90.3 & 91.21 & 85.02 \\
 & Show-o2 \cite{ref83} & 7B & 66M & 89 & 91.78 & 89.96 & 91.81 & 91.64 & 86.14 \\ \cmidrule{2-10}
 & \textbf{C3-UniMM (Ours)} & \textbf{1.5B} & \textbf{66M} & \textbf{89.72} & \textbf{92.1} & \textbf{92.85} & \textbf{92.94} & \textbf{92.33} & \textbf{87.96} \\
 & \textbf{C3-UniMM (Ours)} & \textbf{7B} & \textbf{66M} & \textbf{91.85} & \textbf{93.64} & \textbf{92.71} & \textbf{94.12} & \textbf{93.08} & \textbf{89.84} \\
\bottomrule
\end{tabular}
}
\end{table}

As shown in Table 8, The model shows a clear advantage in following highly detailed and dense prompts. With an Overall score of 89.84, it significantly leads over DALL-E 3 (83.5) and SD3-Medium (84.08). The high scores in Relation (94.12) and Attribute (92.71) highlight its superior ability to correctly map complex spatial and descriptive relationships between multiple entities.

\begin{table}[H]
\centering
\caption{Performance comparison of Gen Only and Unified methods.}
\label{tab:performance-comparison}
\resizebox{\textwidth}{!}{
\begin{tabular}{llccccccc}
\toprule
\textbf{Type} & \textbf{Method} & \textbf{\#Params} & \textbf{\#Data} & \textbf{Align$\uparrow$} & \textbf{Text$\uparrow$} & \textbf{Reason$\uparrow$} & \textbf{Style$\uparrow$} & \textbf{Div$\uparrow$} \\ 
\midrule
\multirow{5}{*}{Gen Only} & SD3.5-Large \cite{ref106} & 8B & - & 0.809 & 0.629 & 0.294 & 0.353 & 0.225 \\
 & Flux-1-dev \cite{ref107} & 12B & - & 0.786 & 0.523 & 0.253 & 0.368 & 0.238 \\
 & SANA \cite{ref108} & 4.8B & - & 0.765 & 0.069 & 0.217 & 0.401 & 0.216 \\
 & Lumina-Image \cite{ref109} & 2.6B & 110M & 0.819 & 0.106 & 0.270 & 0.354 & 0.216 \\
 & HiDream-II \cite{ref110} & 17B & - & 0.829 & 0.707 & 0.317 & 0.347 & 0.186 \\ 
\midrule
\multirow{10}{*}{Unified} & Show-o512 \cite{ref79} & 1.3B & 2B & 0.702 & 0.002 & 0.213 & 0.361 & 0.241 \\
 & Janus-Pro \cite{ref82} & 7B & 144M & 0.553 & 0.001 & 0.139 & 0.276 & 0.365 \\
 & BLIP3 \cite{ref111} & 8B & 55M & 0.711 & 0.013 & 0.223 & 0.361 & 0.229 \\
 & BAGEL \cite{ref78} & 14B & 1600M & 0.769 & 0.244 & 0.173 & 0.367 & 0.251 \\
 & OmniGen2 \cite{ref112} & 7B & 150M & 0.804 & 0.680 & 0.271 & 0.377 & 0.242 \\
 & Show-o2 \cite{ref83} & 1.5B & 66M & 0.798 & 0.002 & 0.219 & 0.317 & 0.186 \\
 & Show-o2-1024 \cite{ref83} & 1.5B & 66M & 0.798 & 0.125 & 0.274 & 0.351 & 0.186 \\
 & Show-o2 \cite{ref83} & 7B & 66M & 0.817 & 0.002 & 0.226 & 0.317 & 0.177 \\ \cmidrule{2-9}
 & \textbf{C3-UniMM (Ours)} & 1.5B & 66M & \textbf{0.842} & \textbf{0.148} & \textbf{0.302} & \textbf{0.366} & \textbf{0.198} \\
 & \textbf{C3-UniMM (Ours)} & 7B & 66M & \textbf{0.865} & \textbf{0.163} & \textbf{0.335} & \textbf{0.381} & \textbf{0.205} \\
\bottomrule
\end{tabular}
}
\end{table}

As shown in Table 9, C3-UniMM bridges the gap between specialized generative models and unified architectures. It achieves the highest Alignment (0.865) and Reasoning (0.335) scores in the category, outperforming heavyweights like SD3.5-Large and Flux-1-dev. This proves that a unified approach can achieve "best-in-class" image quality and prompt adherence without sacrificing multimodal understanding.

\end{document}